\documentclass[letterpaper,11pt]{article}

\usepackage[utf8]{inputenc}
\usepackage[T1]{fontenc}
\usepackage{times}
\usepackage[margin=1in,top=0.45in]{geometry}
\usepackage{microtype}
\usepackage{graphicx}
\usepackage{xcolor}
\usepackage{tcolorbox}
\usepackage{titlesec}
\usepackage[square,numbers,sort&compress]{natbib}
\usepackage[colorlinks=true,linkcolor=cyan,citecolor=cyan,urlcolor=blue]{hyperref}
\usepackage{hyperref}
\usepackage{fontawesome5}

\usepackage{algorithm}
\usepackage{algorithmicx}
\usepackage{algpseudocode}
\usepackage{amsmath}
\usepackage[export]{adjustbox}  % max width= key: shrink figures to fit a column, never enlarge them
\usepackage{amssymb}
\usepackage{amsthm}
\usepackage{array}
\usepackage{booktabs}
\usepackage{calc}
\usepackage{colortbl}
\usepackage{enumitem}
\usepackage{footnote}
\usepackage{listings}
\usepackage{mathtools}
\usepackage{multirow}
\usepackage{newfloat}
\usepackage{placeins}
\usepackage{seqsplit}
\usepackage{subcaption}
\usepackage{tabularx}
\usepackage{url}
\usepackage{xcolor}
\usepackage{xspace}
\usepackage{wrapfig}
\usepackage{lipsum}
\usepackage{tabularx}
\usepackage{cleveref}

\definecolor{strategyExplorer}{HTML}{E76254}
\definecolor{strategyExploiter}{HTML}{B07D2B}
\definecolor{strategyCautiousExplorer}{HTML}{376795}
\definecolor{figblue}{HTML}{1F6FB5}
\definecolor{figred}{HTML}{E5493A}

\newcommand{\method}{\textsc{Suto}\xspace}
\newcommand{\dataset}{SciEvo\xspace}

\definecolor{abstractbg}{RGB}{230,242,250}
\definecolor{abstractborder}{RGB}{200,210,220}
\titleformat{\section}
  {\sffamily\bfseries\Large}{\thesection}{1em}{}
\setlist{leftmargin=5mm}

\renewenvironment{abstract}
{%
  \begin{center}
  \begin{tcolorbox}[
    colback=abstractbg,
    colframe=abstractborder,
    boxrule=0.6pt,
    arc=6pt,
    width=\textwidth,
    left=8pt,right=8pt,top=8pt,bottom=8pt
  ]
}
{%
  \end{tcolorbox}
  \end{center}
}

\newcommand{\papertitle}{Science Utopia? Closed-Loop LLM Simulation of\texorpdfstring{\\}{ }Academic Research Ecosystems}
\newcommand{\paperauthors}{%
  Yiqiao Jin$^{1\ast}$, Yiyang Wang$^{1\ast}$, Lucheng Fu$^1$, Bing He$^1$, Siheng Xiong$^1$, Yijia Xiao$^2$,\\[0.3em]
  B. Aditya Prakash$^1$, Josiah Hester$^1$, Srijan Kumar$^1$, James Evans$^3$, Jindong Wang$^{4\dag}$%
}
\newcommand{\paperaffiliations}{%
  $^1$Georgia Institute of Technology, $^2$University of California, Los Angeles,\\[0.2em]
  $^3$University of Chicago, $^4$William \& Mary%
}
\newcommand{\papercontact}{%
  \texttt{\{yjin328,ywang3420\}@gatech.edu}, \texttt{jdw@wm.edu}%
}
\hypersetup{
  pdftitle={\papertitle},
  pdfauthor={Yiqiao Jin, Yiyang Wang, Lucheng Fu, Bing He, Siheng Xiong, Yijia Xiao, B. Aditya Prakash, Josiah Hester, Srijan Kumar, James Evans, Jindong Wang}
}

\begin{document}
\thispagestyle{empty}

% Institution logos, in affiliation order, with their original aspect ratios.
\noindent
\includegraphics[height=0.65cm,valign=c]{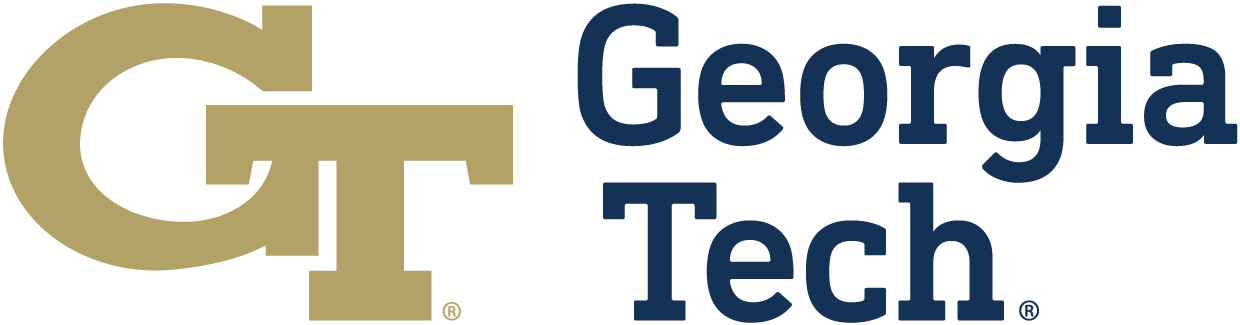}\hfill
\includegraphics[height=0.65cm,valign=c]{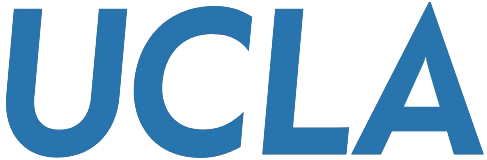}\hfill
\includegraphics[height=0.65cm,valign=c]{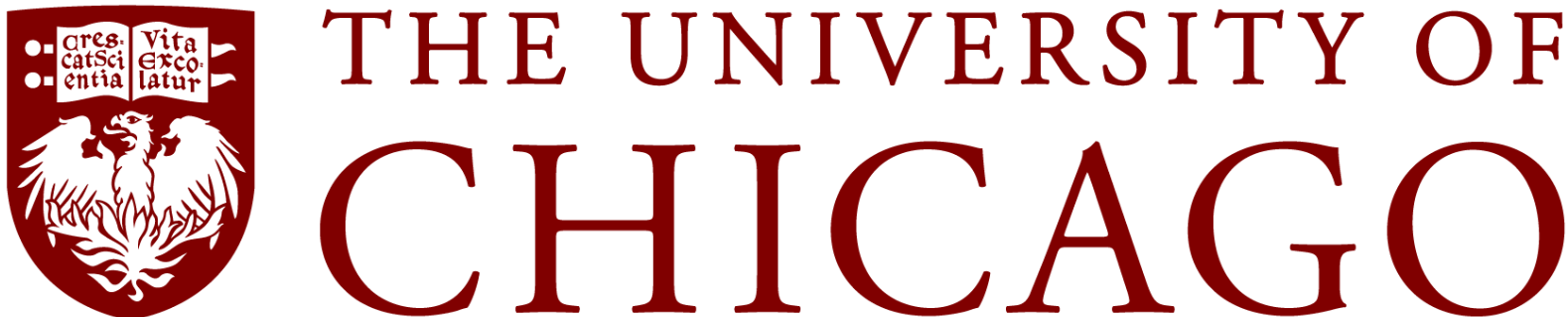}\hfill
\includegraphics[height=0.65cm,valign=c]{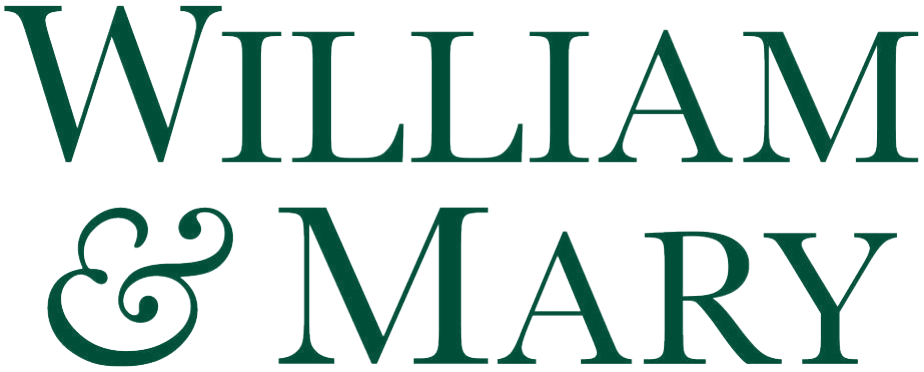}\par
\vspace{0.1cm}
\noindent\rule{\textwidth}{0.8pt}

\begin{center}
  {\sffamily\bfseries\fontsize{16}{20}\selectfont\papertitle\par}
  \vspace{0.4cm}
  {\sffamily\normalsize\paperauthors\par}
  \vspace{0.2cm}
  {\small\paperaffiliations\par}
  \vspace{0.15cm}
  {\small\papercontact\par}
  \faIcon{globe} Website: \url{https://Ahren09.github.io/ScienceUtopia} \\
    \faIcon{github} GitHub: \url{https://github.com/Ahren09/ScienceUtopia}
\end{center}
\begingroup
\renewcommand{\thefootnote}{\fnsymbol{footnote}}
\footnotetext[1]{Equal contribution.}
\footnotetext[2]{Corresponding authors: jdw@wm.edu.}
\endgroup

\begin{abstract}
Scientific progress emerges from a longitudinal ecosystem in which researchers, institutions, funding agencies, collaboration networks, and the scientific literature co-evolve.
% The rapidly-adopted AI makes it promising to perform a comprehensive study of these complicated and evolving aspects.
As AI becomes increasingly involved throughout the scientific research cycle, understanding these interconnected and evolving processes becomes increasingly important. 
% what will happen if every researcher, funding agency, and institution increasingly use AI for all aspects of scientific research?
We introduce \textbf{\method}, a persistent, closed-loop LLM-agent simulation framework for studying academic research ecosystems.
\method models interconnected scientific processes such as research-direction choice, collaboration, submission, peer review, resubmission, citation, funding, and researcher attrition, while maintaining evolving states across simulated years.
% \method enable systematic study through explicit institutional rules, information channels, and seed-controlled stochasticity, supporting matched counterfactual experiments and targeted mechanism interventions. 
% Explicit institutional rules and information channels make \method a controlled testbed for matched counterfactual experiments and targeted mechanism interventions.
Its configurable institutional mechanisms and information channels provide a controlled testbed for matched counterfactual experiments and targeted interventions. 
Across $61$ simulation worlds, \method simulates over $40{,}000$ researchers from $8{,}000$ institutions, producing ${\sim}400{,}000$ publication decisions and $1.2$ million LLM-generated peer reviews.
Using these longitudinal simulations, we find that rejection-driven resubmission substantially amplifies reviewer burden beyond population growth alone, cautious exploration balances citation impact with career success and long-term topic diversity, and resource inequality can emerge even without detectable cumulative advantage from narrowly winning early funding. 
% Our analysis reveals several system-level patterns. Rejection-driven resubmission amplifies reviewer burden beyond the effect of new researcher entry alone. Cautious exploration, i.e., frequent moves into nearby topics, chieves higher citation impact than exploitation while retaining its acceptance and funding advantage. Populations dominated by cautious exploration also sustain greater long-run topic diversity. Barely crossing an early funding threshold does not produce a detectable cumulative funding advantage. Finally, peer-review interventions show that revealing author information shifts review scores, whereas additionally disclosing collaboration-network proximity has little effect. 
% These longitudinal simulation trajectories provide a large-scale resource for studying how researcher behavior, institutional mechanisms, and their interactions shape scientific ecosystems over time. 
Code is available at \href{https://github.com/Ahren09/ScienceUtopia}{github.com/Ahren09/ScienceUtopia}. 
\end{abstract}
\section{Introduction}
% Scientific progress emerges from repeated interactions among researchers, institutions, publication venues, funding, and the evolving scientific literature. 
Artificial intelligence (AI) is rapidly reshaping the scientific research landscape, with large language models (LLMs) increasingly participating in research ideation, experimentation, scientific writing, and peer review~\citep{lu2024ai,yamada2025ai,jin2024agentreview,yu2024researchtown}. 
% Scientific progress emerges from an evolving institutional \emph{ecosystem} shaped by the continuous interplay among researchers, publication venues, funding systems, and the scientific literature. 
As AI adoption extends across interconnected research activities~\citep{li2026can,weng2025cycleresearcher}, % AI is influencing not only  individual discoveries but the organization and evolution of the scientific ecosystem. 
its implications reach beyond individual discoveries to the organization and evolution of the scientific ecosystem. 
% Its trajectory depends not only on individual discoveries, but also on the institutions through which ideas are produced, evaluated, funded, and accumulated over time. 
At the \emph{micro level}, 
AI-mediated decisions can reshape how researchers choose research topics, form collaboration networks, pursue publication, and evaluate submissions, as well as how funding agencies allocate resources. 
% choices about research directions, peer-review judgments, and funding allocation shape individual researchers' incentives, opportunities, and career trajectories. 
At the \emph{macro level}, 
these decisions collectively reshape the direction and organization of science: which ideas gain influence, which research directions receive sustained support, and who remains able to contribute. 
% these decisions accumulate and interact through publication, citation, and resource-allocation processes, potentially altering scientific productivity, research diversity, reviewer workload, and the composition of the research community. 
Crucially, these outcomes feed back into subsequent research, reshaping the knowledge, incentives, and opportunities that guide future decisions.
Such feedback can amplify or counteract the effects of individual choices, so improvements in isolated research activities need not translate into a more productive, diverse, or sustainable scientific ecosystem.
This transformation creates an urgent need for a \emph{closed-loop, longitudinal perspective} to understand not only how AI changes \emph{individual} research activities, but also how these changes propagate over time through an evolving scientific environment to produce \emph{ecosystem-level} benefits and unintended consequences.

\noindent \textbf{Challenges.}
% Recent advances in large language models (LLMs) create new opportunities for studying scientific dynamics. As LLMs increasingly participate in scientific writing, reviewing, and decision-making, 
% LLM agents provide a natural substrate for controlled studies of scientific processes that are themselves becoming increasingly AI-mediated (Appendix~\ref{app:discussion}). 
Tracing these effects and causal pathways, however, is difficult in real academia, where controlled institutional interventions are often infeasible and counterfactual outcomes under alternative policies are unobservable. 
% Simulation offers a controlled setting for testing institutional mechanisms, with LLM agents providing a natural substrate for studying increasingly AI-mediated scientific processes (Appendix~\ref{app:discussion}). 
% Realistically capture the \emph{co-evolution of researcher behavior and the scientific environment} while retaining control over the institutional mechanisms under study remains challenging. 
The researcher population, available resources, collaboration networks, and scientific record evolve with prior decisions and, in turn, shape subsequent research.
Although LLM agents provide a natural substrate for studying increasingly AI-mediated scientific processes in controlled settings through simulation (Appendix~\ref{app:discussion}), faithfully capturing this co-evolution poses three challenges. 
First, \emph{scientific fidelity} requires more than plausible agent responses: research choices and evaluations must be grounded in scientific content and reflect differences in researchers' expertise and strategies. 
Second, \emph{closed-loop simulation} requires research outcomes to reshape the conditions for subsequent research. 
Lower acceptance rates may increase resubmissions, consuming authors' resources and adding review demand. 
Researchers who exhaust their budgets may leave, reducing both future submissions and the reviewer pool. 
Accepted papers expand the citable literature and update the publication records used in funding decisions. These changes must carry forward into the next year's decisions, so that both researcher behavior and the community's capacity to sustain research evolve within the simulation. 
Third, maintaining effective \emph{experimental control} requires separating the mechanism being tested from the changes it produces, whereas ineffective design would suppress the feedback effects the simulation is intended to study. 
% Scientific processes are deeply interconnected, with earlier outcomes shaping later decisions.
% Researchers' expertise, resources, reputation, collaborations, and publication records persist across years. 

% Agents must interact with papers, topics, venues, and citation relations while accessing only information available at each simulated point in time.
% Existing scientific corpora are largely designed for retrieval or retrospective analysis, rather than temporally constrained, simulation-ready access to these interconnected components. 
% =====

% Evaluation must therefore examine both local and downstream effects through controlled interventions. 
% Its validity must instead be assessed through behavioral checks, controlled interventions, matched counterfactual worlds, and mechanism tracing, while carefully distinguishing simulated dynamics from claims about human scientific behavior. 
% Without explicitly modeling these feedback loops, it remains difficult to determine whether an institutional intervention produces sustained improvements in scientific incentives and outcomes or instead redistributes costs, constraints, and pressures across the broader system. 
% In the real world, controlled interventions are costly and slow, scientific actors and institutions co-evolve, and consequential counterfactuals such as outcomes under alternative funding or review policies are fundamentally unobservable. 

\begin{figure*}[t!]
\centering
\includegraphics[width=0.98\linewidth]{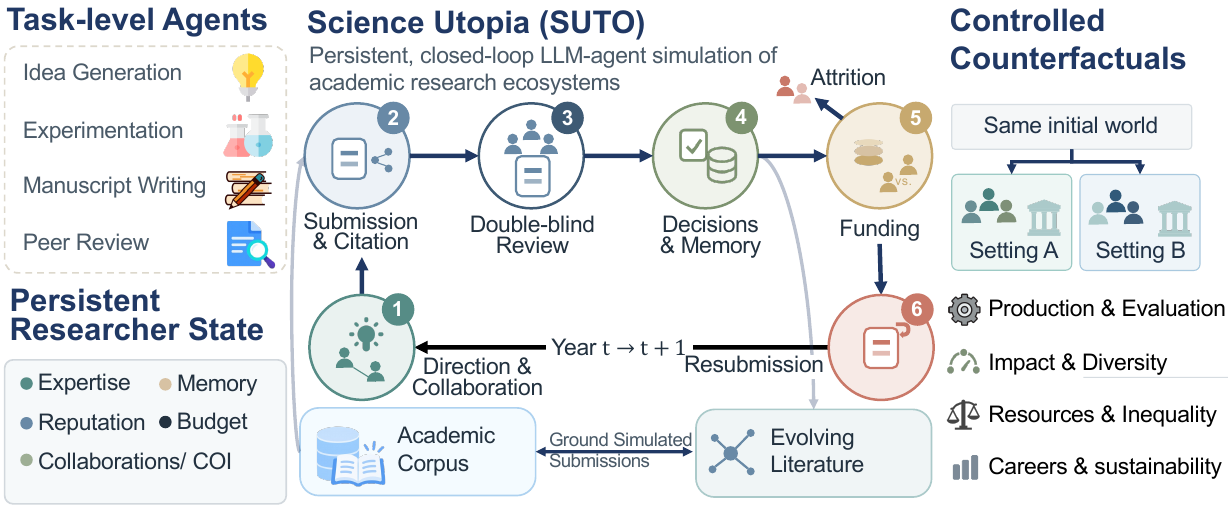}
\vspace{-.1in}
\caption{
\method simulates each year as a six-phase cycle whose outcomes update persistent researcher states and inform subsequent decisions.
}
\label{fig:teaser}
\vspace{-3mm}
\end{figure*}

\noindent \textbf{This Work.}
We introduce \method (\textbf{\underline{S}}cience \textbf{\underline{Uto}}pia), a persistent, closed-loop LLM simulation framework for the longitudinal study of research ecosystems. 
\method integrates research-direction choice, collaboration, submission, peer review, resubmission, citation, funding, and researcher attrition through persistent researcher states and a shared scientific record.
This structure supports matched counterfactual experiments that vary an institutional mechanism and follow its downstream effects under the same simulation rules.
We investigate three questions. 
% First, \textit{how do local institutional rules generate delayed system-level effects?}
% Holistic studies of researcher population growth and rejection recycling show that resubmission creates a persistent manuscript backlog, amplifying submission growth and reviewer burden.
First, \emph{how do publication and funding rules shape the long-term dynamics of research ecosystems?} 
% Resubmission creates a persistent backlog, amplifying submission growth and reviewer burden beyond the effects of researcher population growth alone. Stricter acceptance criteria intensify this burden by shifting rejected manuscripts into future review rounds. Higher research output under fixed funding creates another resource pressure: publication volume increases, but researcher attrition also accelerates. 
Second, \textit{how do individual research strategies shape career outcomes and the diversity of science?}
% Cautious explorers, who pursue new topics close to their existing expertise, retain much of exploiters' acceptance and funding advantage while achieving greater citation impact. Populations dominated by cautious explorers also sustain greater long-run topic diversity than those dominated by either explorers or exploiters, showing that larger individual topical departures do not necessarily produce a more diverse research ecosystem. 
Third, \textit{under what conditions do cumulative advantage and bias emerge in funding and peer review?} 
%  \textit{does narrowly winning early funding lead to cumulative advantage in later outcomes?}
% Narrowly winning early funding produces no significant increase in subsequent funding. In peer review, revealing author identity shifts scores more than additionally disclosing collaboration-network proximity, distinguishing the effects of knowing who an author is from knowing how closely connected they are to the reviewer. 
Our analyses span $61$ simulation worlds containing over $40{,}000$ researcher agents across $8{,}000$ institutions and approximately $390{,}000$ simulated researcher-years.
Overall, \method generates about $400{,}000$ publication decisions and $1.2$ million peer reviews. % (Table~\ref{tab:dataset-stats})
Our findings offer insights for future academic systems: 
\begin{itemize}[leftmargin=*]
    \setlength\itemsep{0em}
    \item \textbf{Repeated resubmission sustains a significantly greater manuscript backlog and substantially amplifies per-reviewer workload compared with researcher population growth.} % , whereas population growth alone leaves per-reviewer workload close to baseline.
    Population growth alone expands the reviewer pool alongside submissions, leaving review demand near baseline (2.07 vs. 2.25 reviews per active reviewer). Resubmission alone raises this demand to 6.64. 
    Stricter acceptance criteria compound this burden by shifting review work into future rounds, inducing 14\% more review workload, even when the volume of new submissions changes little. 

    \item \textbf{Publication growth can mask declining scientific participation.} 
    Doubling the submission limit per project increases accepted output by $63$--$69\%$ in university-only worlds, yet reduces long-term participation even when funding capacity scales with demand. Under fixed grant capacity, the share of initial researchers active at year ten falls from $39\%$ to $24\%$. 
    % This pattern persists across university-only and mixed university--industry populations, indicating that productivity gains can increase exclusion when institutional resources do not scale with output.

    % \item \textbf{Cautious exploration balances career rewards, citation impact, and scientific diversity.}
    \item \textbf{Exploring novel but related topics balances citation impact with publication and funding success.}
    Researchers exploring new topics close to expertise (\emph{cautious explorers}) achieve greater citation impact than those extending prior work (\emph{exploiters}) and those moving into distant areas (\emph{explorers}), with only a modest decline in paper acceptance and funding. 
    When most researchers follow this strategy, long-term research activities remain broadly distributed across topics over time. % when most either continue established lines of work or move into distant areas.
    % Populations dominated by cautious explorers also sustain greater long-run topic diversity than those dominated by either explorers or exploiters, showing that larger individual topical departures do not necessarily produce greater collective diversity.

    \item \textbf{Narrow early funding wins show no significant career advantage over the subsequent three years.} 
    However, resources become concentrated over time and nearly half of researchers exhaust their resources.

    \item \textbf{Revealing author information modestly raises review scores.} % while disclosing collaboration ties adds little.
    Revealing the first author's pseudonymous name and institution raises scores by $0.055$--$0.077$ points relative to anonymous review on the $1$--$5$ scale.
    Disclosing shared collaborators or other network connections has little further effect once author information is visible.
    % Once this information is visible, additionally disclosing collaboration-network proximity has a smaller effect, distinguishing the influence of author information from that of explicitly disclosed professional connections.
\end{itemize}

% We hope this work can provide useful insights to the future study and reshape of the research ecosystem; more importantly, to better cope with the AI-enabled scientific research in today's world.
We hope this work will lay a foundation for understanding and reshaping research ecosystems and, more importantly, help shape the future of scientific research in the age of AI. % as AI transforms how knowledge is created, evaluated, and shared.

% \begin{itemize}[leftmargin=*]
% \setlength\itemsep{0em}
%     \item \textbf{Framework.}
%     We introduce \method, a persistent, closed-loop LLM-agent framework that models research-direction choice, collaboration, submission, peer review, resubmission, citation, funding, and attrition within a single longitudinal ecosystem.
%     % Researcher states and institutional outcomes persist across years, enabling local decisions to influence subsequent scientific behavior and system-level dynamics.
   
%     \item \textbf{Evaluation \& Insights.}
%     We conduct large-scale evaluations using seed-controlled, matched counterfactual worlds and mechanism-specific interventions.
%     Our studies show that rejection recycling amplifies researcher-driven submission growth and disproportionately increases reviewer burden; cautious exploration balances institutional reward with scientific impact and sustains greater ecosystem diversity; and controlled null results reveal why hypothesized funding and review effects do not arise automatically when the corresponding information or feedback channels are absent.

% \end{itemize}

\section{The \method Framework}
\label{sec:method}

% As shown in \Cref{fig:framework}, 

As shown in Appendix~\Cref{fig:framework}, \method simulates academic research as an LLM-powered, closed-loop cycle, with research, peer review, publication, and funding outcomes shaping subsequent decisions across simulated years.

%  experiment-specific configurations appear in Appendix~\ref{sec:eval-setup}.

\vspace{-1mm}
\subsection{Researchers, Institutions, and Venues} 
\label{sec:participants}
\vspace{-1mm}
% Researchers are the persistent units of the simulation. 
\noindent \textbf{Researchers.}
A researcher $i$ maintains a state in year $t$, $s_i^{t} = \bigl(E_i,\rho_i^{t},b_i^{t},M_i^{t},C_i^{t}\bigr)$, where $E_i$, $\rho_i^{t}$, and $b_i^{t}$ are the researcher's expertise topics, reputation, and available budget, respectively.  
$C_i^{t}$ is the set of conflicts of interest induced by affiliation and co-authorship. 
$M_i^{t}$ is the agent's memory that records previous events, such as publications, peer-review experience, and funding outcomes.
The memory connects past outcomes to subsequent decisions: prior reviews can guide revision and resubmission, while publication outcomes can influence whether a researcher continues or changes research direction. 
This feedback enables us to study how institutional rules shape research trajectories over time beyond their immediate effects on acceptance and funding. 
The budget $b_i^{t}$ links research and funding outcomes to continued participation in the ecosystem. 
% determines whether a researcher can continue participating in the ecosystem, creating a resource constraint that connects short-term research and funding outcomes to longer-term participation and attrition. 
At the end of each year $t$, it is updated as $b_i^{t+1} = b_i^{t} + f_i^{t} - \sum_{a\in\mathcal{A}_i^{t}} \mathrm{cost}(a)$,
where $f_i^{t}$ is funding received during the year, $\mathcal{A}_i^{t}$ is the set of actions taken by researcher $i$, and $\mathrm{cost}(a)$ is the cost of action $a$. 
Researchers whose budget is exhausted become inactive and leave the active population. 
The budget can be interpreted as an abstract resource constraint, analogous to AI-usage tokens. 
% The budget can be thought of as ``tokens'' for AI usage today.

To examine how funding arrangements shape researchers' strategies and  participation, we distinguish two institutional settings, motivated by differences between academic and industrial science.  
\emph{University researchers} rely on external competitive grants, whereas \emph{industry researchers} receive internal funding tied to accepted papers. 
Meanwhile, strategies between exploiting familiar topics and exploring new ones shape both individual career outcomes and the collective development of science~\citep{foster2015tradition}.  
Prior work characterizes this trade-off along two dimensions: \emph{exploration propensity}, how often researchers enter unfamiliar areas, and \emph{exploration distance}, how far they depart from their prior work~\citep{yang2023cautious}.
To examine how institutional rules reward different \emph{exploration strategies}, we consider three representative approaches: 
\emph{explorers} favor distant moves into new areas; 
\emph{cautious explorers} frequently explore nearby areas, balancing novelty with continuity in prior expertise; 
and \emph{exploiters} preferentially continue along established research directions. 
% \emph{explorer}, \emph{exploiter}, or \emph{cautious explorer}, which determine the degree of departure from a researcher's existing expertise and prior research trajectory. 

\noindent \textbf{Funding Programs.}
\method includes two families of funding programs.
Foundation-oriented programs prioritize theoretical and foundational work, while mission-oriented programs favor application-driven areas such as autonomous systems and security.
This distinction is motivated by
research-support and mission-focused programs such as NSF CAREER and DOE's Genesis Mission~\citep{nsf_career,doe,braun1998role,mazzucato2018mission}.
Each program specifies eligible topics, evaluates submitted applications, and funds the top-ranked fraction of applications assigned to its panel.

\noindent \textbf{Venues and scientific artifacts.}
Scientific artifacts are grounded in real literature so that ecosystem dynamics are not dominated by the factual reliability of fully generated papers.
Each conference $c$ covers a subset of research topics and has an acceptance rate $\alpha_c$.
A submission must be topically compatible with its venue.
Given its selected research direction, an agent retrieves temporally valid papers from the corpus and selects top papers for submission to a compatible venue.
This retrieval-grounded design provides scientific material for review and citation while restricting retrieved artifacts by year.
By default, each completed project allows at most one new submission, which may cite previously published work, including the researcher's own papers.
Accepted papers and citations are tracked, making the evolving literature available for citation in subsequent years. 

\vspace{-1mm}
\subsection{Simulation Pipeline}
\vspace{-1mm}
Each simulated year follows the same execution order.
Simulated year $t$ is mapped to a real calendar year, so agents can only submit or cite papers that would have been available by that point in time.
The phases below describe the default loop. Individual experiments may invoke additional controlled interventions without changing the pipeline. The motivation is in Appendix~\ref{app:closed-loop-motivation}.

\noindent \textbf{Phase 1: Research Direction and Collaboration.}
Active agents whose previous projects have concluded (if any) select a new research direction $d\in\mathcal{D}$, where $\mathcal{D}$ is the fixed set of available research directions, based on their expertise, resources, and memory.
A direction $d$ lasts for $T_d$ simulated years and costs $\kappa$ budget units per active year, so its total cost is $\kappa T_d$.
This duration makes resource pressure visible: longer directions consume more budget before producing new outputs.
Agents can choose to form co-authorship ties when collaboration is enabled. 
Agent $i$ scores a candidate collaborator $j$ using $\psi(i,j)=\sum_m w_m\,\phi_m(i,j),$ where each feature $\phi_m(i,j)$ captures one collaboration signal, such as topical similarity, recent-publication overlap, preferential attachment, or triadic closure, and $w_m$ is its weight.

\noindent \textbf{Phase 2: Paper Submission and Citation.}
Authors may submit various numbers of new papers in a year depending on their bandwidth.
For each potential submission, \method retrieves candidate arXiv papers from \dataset that match the agent's direction and are available in the current simulated year; the agent then chooses one candidate artifact and a compatible venue.
Research costs and resubmission fees enter the budget update.
After choosing the artifact, the agent cites prior papers from the accessible literature. Self-citation behavior is analyzed in Appendix~\ref{app:self-citation}. 

\noindent \textbf{Phase 3: Peer Review.}
Each submission is assigned three reviewers $R_p$. Each reviewer $j$ uses the title, abstract, and topics to provide a recommendation $u_{jp}$ and written justification for paper $p$, following prior work~\citep{jin2024agentreview}.
Reviews are double-blind by default. 
% who produce an overall recommendation $u_{jp}$ for reviewer $j$ on paper $p$, together with a written justification, following standard review procedures and previous work~\citep{jin2024agentreview}.

\noindent \textbf{Phase 4: Paper Decisions and Memory Update.}
For a paper $p$ reviewed by panel $R_p$, the decision score is the mean recommendation $\bar u_p=\frac{1}{|R_p|}\sum_{j\in R_p}u_{jp}.$
Within each year, conference $c$ ranks its submissions $S_c$ by $\bar u_p$ and, for nonempty $S_c$, accepts the top $\max(1,\operatorname{round}(\alpha_c |S_c|))$ papers, where $\alpha_c$ is the acceptance rate.
Authors retain experiences such as  reviews and publication decisions in memory, which shape future behavior. 
For example, negative reviews may affect resubmission decisions and prompt harsher evaluations of others' work, while publication success may reinforce a research direction.
Accepted papers enter the citable literature and can be cited by submissions in later years.

\noindent \textbf{Phase 5: Funding Allocation and Attrition.}
University researchers submit short proposals grounded in their expertise and prior papers. 
Funding agents evaluate applications for program fit, research potential, and track record, considering both accepted and rejected papers.\footnote{We do not simulate the ``proposal-funding'' pipeline by assuming that the publications are positively associated with peer-review evaluations and funding outcomes of research proposals \citep{cabezas2013reviewers}.}
Under rate-based funding, each panel ranks its $n$ applications and funds the top $k=\max(1,\lfloor r n\rfloor)$ applicants, where $r$ is the program's funding rate.
To study how \emph{funding conservatism} shapes downstream research trajectories, we introduce a controlled intervention that penalizes topical departure while leaving the underlying evaluation unchanged. 
We convert researcher $i$'s within-panel rank into a normalized score $e_i\in[0,1]$, with higher scores indicating better ranks.
We measure topical departure by $q_i\in[0,1]$, the percentile rank of the researcher's mean paper distance among the given year's applicants. % with scored papers. 
We then adjust the funding score as $\tilde e_i=e_i-\lambda q_i$, where $\lambda\ge0$ controls the penalty strength, and award funding to applicants with the highest adjusted scores. 
Using normalized ranks makes the penalty independent of the raw score scales. 
The reported experiments use $\lambda=0$, preserving the original ranking.

\noindent \textbf{Phase 6: Resubmission.}
From the second year onward, agents with previously rejected papers first decide whether to abandon, revise, or resubmit those papers.
The decision is conditioned on the received reviews, the paper's prior submission history, the agent's current resources, and the resubmission cost.
Resubmitted papers re-enter the review process alongside new submissions.

\vspace{-1mm}
\section{Experiments and Findings}
\label{sec:evaluation}
\vspace{-1mm}

% We use \method to answer three research questions that are difficult to study in real academia.
% First, do local institutional rules create system-level consequences through feedback loops (\S\ref{sec:results-feedback})?
% Second, how do individual research incentives connect to ecosystem-level outcomes (\S\ref{sec:results-incentives})?
% Third, can controlled interventions identify which mechanisms and information channels generate, or fail to generate, commonly hypothesized patterns (\S\ref{sec:results-counterfactual})?
% Experimental configurations and metrics are in Appendix~\ref{sec:eval-setup}.

We organize our findings across three levels of the scientific ecosystem: researchers, examining how research strategies shape careers (\S\ref{sec:results-human}); research outcomes, connecting topical departure to impact and disruption and examining how resubmission and peer review shape publication outcomes (\S\ref{sec:results-research}); and institutional mechanisms, examining how funding capacity and allocation shape resource access and continued participation (\S\ref{sec:results-institution}).
Finally, we examine the ecosystem-level dynamics emerging over time (\S\ref{sec:results-overall}). Experimental details are in Appendix~\ref{sec:eval-setup}. 

\vspace{-1mm}
\subsection{Human Factors: Exploration Strategies and Careers}
\label{sec:results-human}
\vspace{-1mm}
% \textbf{Cautious Exploration Balances Novelty and Career Reward}
% \label{sec:results-exploration}

\begin{figure*}
\centering
\includegraphics[width=0.98\linewidth]{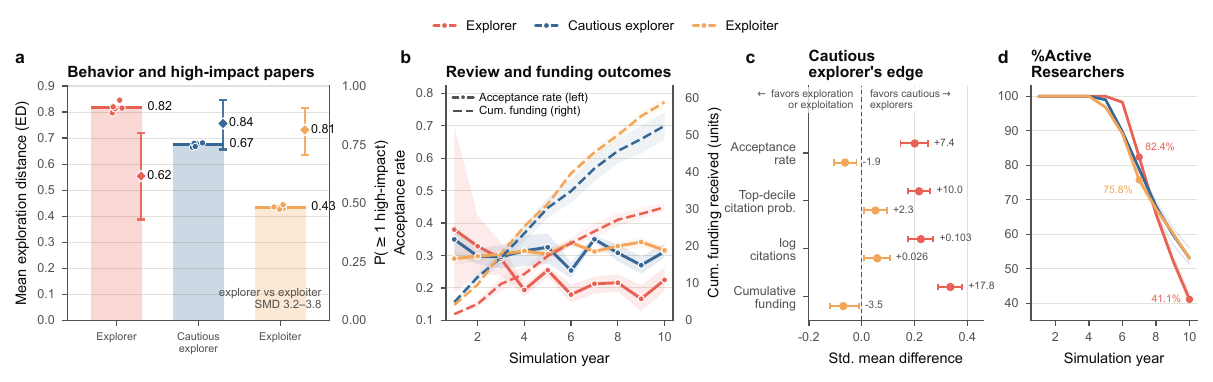}
\vspace{-2mm}
\caption{
\textbf{Career outcomes across research strategies, including \textcolor{strategyExplorer}{explorers}, \textcolor{strategyExploiter}{exploiters}, and \textcolor{strategyCautiousExplorer}{cautious explorers}.}
% \textcolor{strategyExplorer}{Explorers} make the largest topical departures,
% \textcolor{strategyExploiter}{exploiters} remain closest to prior expertise, and
% \textcolor{strategyCautiousExplorer}{cautious explorers} balance the two.
\textbf{(a)} Exploration distance and high-impact-paper probability across five calibration worlds. % Mean exploration distance (left axis, bars and per-world dots) confirms that the three strategies induce distinct research behaviors. Cautious explorers are most likely to produce at least one high-impact paper (right axis, diamonds with min--max range across worlds), followed by exploiters and explorers.
\textbf{(b)} Acceptance and cumulative funding in the same calibration worlds. % Explorers face the lowest acceptance rates (left axis, solid lines). This disadvantage compounds into substantially lower cumulative funding over the decade (right axis, dashed lines).
\textbf{(c)} Cautious-explorer contrasts in three $5{,}000$-researcher confirmatory worlds: standardized mean differences with $95\%$ CIs. Here citation probability uses researcher-level citation totals. % (standardized mean differences with $95\%$ CIs; here color marks the comparison strategy the cautious explorer is measured against).
\textbf{(d)} Active researchers in the confirmatory worlds. % Explorers initially remain active at higher rates but experience the sharpest late-stage attrition, falling to $41.1\%$ active by year 10 compared with approximately $53\%$ for the other strategies.
}
\label{fig:exploration_strategies}
\vspace{-4mm}
\end{figure*}

% Scientists face a fundamental trade-off between exploiting familiar research areas and exploring new ones~\citep{foster2015tradition}.  
% Recent work shows that this trade-off is inherently two-dimensional: researchers
% differ in \emph{how often} they move into unfamiliar areas
% (\emph{exploration propensity}) and \emph{how far} those moves depart from
% their prior work (\emph{exploration distance})~\citep{yang2023cautious}. 
% This distinction identifies a particularly interesting pattern,
% \emph{cautious exploration}, through relatively short moves
% in knowledge space, balancing novelty with continuity in prior expertise. 
% Motivated by this characterization, we compare three canonical research strategies~\citep{yang2023cautious}: \emph{explorers}, which favor distant moves into new areas; \emph{cautious explorers}, which explore new areas while remaining close to their existing expertise; and \emph{exploiters}, which preferentially continue along established research directions. 
We track \emph{explorers}, \emph{exploiters}, and \emph{cautious explorers} (\S\ref{sec:participants}) over $10$ years while holding initial funding and reputation constant. 
We use two complementary distances to quantify researchers' movement through the knowledge space~\citep{huang2022revisiting,yang2023cautious,zhang2024gender}.
For paper $p$ initially submitted by researcher $i$ in year $t$, we compute its \emph{paper distance} from researcher $i$'s recent work as
$\nu_{p,i}^{t} = 1-\cos(\mathbf{p},\mathbf{a}_i^{t}) \in [0,2]$, where $\mathbf{p}$ is the paper embedding and $\mathbf{a}_i^{t}$ is the mean embedding of the author's work over the preceding four years. 
A researcher's \emph{exploration distance (ED)} is the mean annual cosine distance from each paper to their most recent paper.
Thus, paper distance measures departure from recent expertise, whereas ED measures successive movement through knowledge space.  
The resulting trajectories match the intended strategies: explorers have the largest ED, exploiters the smallest, and cautious explorers lie between them (Figure~\ref{fig:exploration_strategies}a). 
% We measure citation impact at a fixed paper age, counting citations received within the first $k$ years after publication and including only papers with a complete $k$-year observation window.
% The mean embedding of the author's papers from the preceding three simulated years (before an author's first paper, the mean embedding of their initial expertise-direction descriptions).
% Because cosine similarity between these text embeddings effectively lies in $[0,1]$, novelty runs from ${\sim}0$ (a direct continuation of the author's recent work) to ${\sim}1$ (essentially unrelated to it); a researcher's \emph{career distance} is the mean novelty of their papers.
% Citation impact is always compared at a fixed \emph{paper age}: ``citations by age $k$'' counts the citations a paper accumulates within $k$ simulated years of publication, restricted to papers published early enough to have a complete $k$-year window, which avoids comparing the cumulative citations of old and young papers.

\noindent \textbf{Cautious exploration balances novelty with career reward.} % Achieves a Better Balance Between Novelty and Career Reward
Distinct exploration strategies lead to different career outcomes. 
% Exploratory work encounters the strongest institutional barrier at publication: 
Explorers have the lowest acceptance rate ($29.8\%$), compared with $37.2\%$ for cautious explorers and $39.2\%$ for exploiters (Figure~\ref{fig:exploration_strategies}b), and are 
also less likely to produce a high-impact paper ($0.62$ versus $0.84$ and $0.81$ in Figure~\ref{fig:exploration_strategies}a). 
Cautious explorers outperform explorers in acceptance, citation outcomes, and cumulative funding (Figure~\ref{fig:exploration_strategies}c), while relative to exploiters they trade modestly lower acceptance and funding for stronger citation impact. 
Thus, exploration is most effective when it extends existing expertise without losing institutional support. 

\noindent \textbf{The costs of distant exploration accumulate over time.}
% Explorers initially remain active at higher rates, but this pattern reverses as publication and funding outcomes accumulate. 
Explorers' early advantage in remaining active reverses as publication and funding outcomes accumulate. 
By year $10$, only $41.1\%$ of explorers remain active, compared with ${\sim}53\%$ of cautious explorers and exploiters. 
% The risks of distant exploration are not immediately visible, but emerge through long-run career dynamics. 
% Explorers initially remain active at the highest rate: through year seven, $82.4\%$ are still active, compared with $75.8\%$ of exploiters (Figure~\ref{fig:exploration_strategies}d). 
% Their longer research cycles sustain research activity before publication and funding outcomes fully accumulate. 
% Over time, however, their lower rate of converting projects into research outcomes reduces opportunities to accumulate citations, reputation, and subsequent funding. 
% By year 10, explorer survival falls to $41.1\%$, compared with $53\%$ for alternative strategies. 
% This disadvantage extends to the highest-impact outcomes: only 2 explorer papers reach the 99th citation percentile, compared with roughly 4 for the other strategies (Figure~\ref{fig:paper_novelty_scores}). 
This delayed reversal suggests that early persistence can mask accumulating disadvantages in publication and funding. 

\begin{figure*}[t]
\centering
\includegraphics[width=0.98\linewidth]{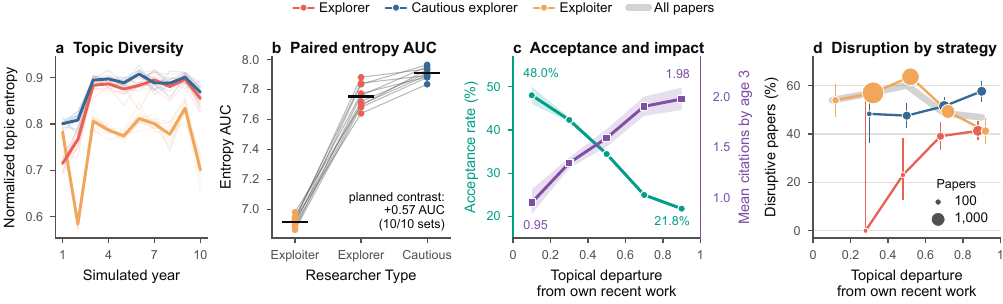}
\vspace{-3mm}
\caption{
\textbf{(a)} Normalized topic entropy over ten simulated years.
\textbf{(b)} Ten-year entropy AUC (Eq.~\ref{eq:topic-entropy}) across matched counterfactual worlds; lines connect worlds within each matched set.
\textbf{(c)} Acceptance rate and mean citations by age three across topical-departure bins.
\textbf{(d)} Share of disruptive papers (CD $>0$ at age three) by strategy and topical departure; marker area denotes the number of papers and gray pools all strategies.
(a, b) compare ten matched sets of explorer-heavy, exploiter-heavy, and cautious-heavy worlds.
(c, d) pool papers from three large-scale ten-year simulation worlds. 
% \textbf{Research strategy shapes ecosystem diversity (a, b). Greater research shift faces a review penalty despite higher downstream impact (c) and raises disruption only for cautious explorers (d).} We compare three matched counterfactual settings that differ only in their population mix: explorer-heavy, exploiter-heavy, or cautious-heavy. Each setting contains ten matched worlds that share institutions, venues, and corpus. Colors mark the dominant strategy of each world in (a, b) and the author's strategy in (d).
% \textbf{(a)} Topic diversity over ten simulated years. Exploiter-heavy worlds rapidly concentrate on fewer directions, whereas cautious-heavy worlds maintain the highest diversity. 
% \textbf{(b)} Ten-year diversity AUC (Eq.~\ref{eq:topic-entropy}) for each matched world. Cautious-heavy worlds are more diverse than both alternatives in all ten matched comparisons. 
% \textbf{(c)} Papers from the three confirmatory worlds are grouped into five equal-width bins measuring research-shift.
% Left axis: submission acceptance rates % with Wilson $95\%$ CIs
% decline from $48.0\%$ in the lowest-shift bin to $21.8\%$ in the highest.
% Right axis: among accepted papers, mean citations by year three increase from $0.95$ to $1.98$ as research shift rises.
% \textbf{(d)} Share of disruptive papers (CD $>0$ at age three) in the same bins, for the $8{,}507$ papers with a defined CD index. Colored lines show strategies with Wilson 95\% confidence intervals, and marker area grows with the number of papers in each bin. The gray line pools all papers.
}
\label{fig:topic_diversity}
\vspace{-3mm}
\end{figure*}

\noindent \textbf{Cautious exploration sustains long-run scientific diversity.}
To test whether individual-level advantages scale to the ecosystem, we construct matched counterfactual worlds that differ only in researcher composition---explorer-heavy, exploiter-heavy, or cautious-heavy---while holding institutions, venues, and all other simulation settings fixed. 
We measure ecosystem diversity using normalized topic entropy and summarize over the full simulation by the area under the entropy curve: 
\begin{equation}
H_t = -\frac{\sum_{d\in\mathcal{D}} p_{t,d}\log p_{t,d}}{\log |\mathcal{D}|}\in[0,1], \qquad\mathrm{AUC}_H = \sum_{t=1}^{T-1} \frac{H_t + H_{t+1}}{2} \in [0,\, T-1],
\label{eq:topic-entropy}
\end{equation}
Here, $p_{t,d}$ is the fraction of all direction choices made in year $t$ that select $d$, with $\sum_{d\in\mathcal{D}}p_{t,d}=1$. % The denominator uses all $|\mathcal{D}|=53$ available directions, including those not selected that year.
Higher $H_t$ indicates a more even distribution of research exploration across topics. With one-year spacing between observations, $\mathrm{AUC}_H$ summarizes diversity over years $1$ through $T=10$.
% Ecosystem diversity depends not only on how far individual researchers explore, but also on whether new research directions can be sustained over time. 
Varying the proportions of the three strategies substantially changes the resulting research landscape (Figure~\ref{fig:topic_diversity}). 
Cautious-heavy populations maintain $0.11$ higher normalized topic entropy than exploiter-heavy populations and end with $8.5$ more active research directions across all ten matched counterfactual sets (Figure~\ref{fig:topic_diversity}). 
They also outperform explorer-heavy populations in ten-year $\mathrm{AUC}_H$ by $0.156$ on average, despite explorers making larger individual departures from prior expertise. 
Explorer-heavy worlds initially spread research directions more broadly, but manifest greater long-run attrition and lower diversity.
Policies that encourage exploration should therefore also help exploratory researchers remain active. Otherwise, research may gradually concentrate on topics with safer career prospects, leaving fewer people to pursue risky but potentially valuable directions. 
% Research Shift or Topic Departure

\vspace{-2mm}
\subsection{Research Outcomes: Impact, Resubmission, and Peer Review}
\label{sec:results-research} 
\vspace{-2mm}
The researcher-level results raise a natural question: is exploratory work less influential, or simply harder to publish? We first relate topical departure to acceptance, impact, and disruption, then examine how resubmission and reviewer knowledge shape publication outcomes and review workload.

\noindent \textbf{Greater topical departure faces a stronger publication gate but higher impact once accepted.}
Papers that depart further from an author's recent research are substantially harder to publish (Figure~\ref{fig:topic_diversity}c). 
From the lowest to the highest departure bin, acceptance falls from $48.0\%$ to $21.8\%$, while mean review scores decrease from $2.45$ to $2.28$. 
Among accepted papers, however, downstream impact moves in the opposite direction. Mean citations by age three increase from $0.95$ to $1.98$, and the $90$th percentile increases from $3$ to $5$. 
These results reveal a selection--impact trade-off: work farther from an author's recent trajectory is less likely to pass the publication gate, but the subset that is accepted receives greater subsequent impact.
% Thus, a large research shift is risky not because shifted papers have less impact, but because they are less likely to pass review. 
% This helps explain the explorer penalty in Figure~\ref{fig:exploration_strategies}: explorers accumulate more rejections, whereas cautious explorers remain close enough to established directions to pass the review gate while still capturing the impact premium of topical departure (Figure~\ref{fig:paper_novelty_scores}).

\begin{figure*}
\centering
\includegraphics[width=0.98\linewidth]{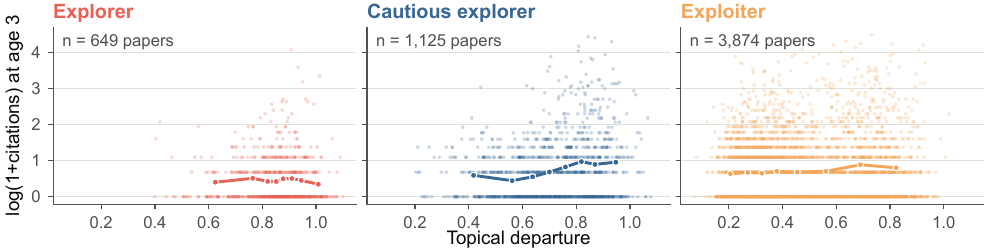}
\vspace{-.1in}
\caption{
\textbf{Topical departure is associated with the clearest citation gains for cautious explorers.} 
Each point represents an accepted paper, with topical departure from the author's recent work on the $x$-axis and citations within three years on the $y$-axis ($\log(1+\text{citations})$). Lines show mean citations within departure octiles.
Citation gains rise most consistently with departure for cautious explorers, while the relationship is weak for explorers and exploiters. 
% \textbf{Research shift pays off mainly for the cautious explorer.} Each point is a paper: its research-shift score ($x$) against its citations by age three ($y$, on a $\log(1{+}\text{citations})$ scale); the connected line traces the mean within each shift octile, one panel per strategy. The citation payoff to topical departure rises most clearly for cautious explorers, whereas for explorers and exploiters a greater shift brings little citation gain---departure is rewarded when it stays close enough to the author's established expertise.
}
\label{fig:paper_novelty_scores}
\vspace{-.1in}
\end{figure*}

% \textbf{The novelty penalty is a selection effect, not a quality effect.}
% Where exactly does the explorer's disadvantage arise---are novel papers worse, or merely filtered?
% Separating the review gate from post-acceptance impact gives a clean answer (
% As shown inFigure~\ref{fig:topic_diversity}c), acceptance falls monotonically with paper novelty, from $48.0\%$ in the least-novel bin to $21.8\%$ in the most novel, and mean review scores decline in parallel ($2.45$ to $2.28$).
% Conditional on acceptance, the gradient reverses: citations by paper age three (i.e., within three simulated years of publication; we fix paper age to avoid comparing the cumulative citations of old and young papers) more than double across the same novelty bins, from $0.95$ to $1.98$, while the 90th percentile rises from $3$ to $5$.
% The ``high-risk, high-reward'' payoff to novelty therefore exists in the simulated citation economy---but only downstream of a review gate that admits novel work at roughly half the rate.
% The explorer penalty of Figure~\ref{fig:exploration_strategies} is thus a selection effect compounded over a career, not evidence that exploratory work is less valuable once published; the strategy whose novelty stays near the gate's tolerance (the cautious explorer) is precisely the one that captures the conditional premium (Figure~\ref{fig:paper_novelty_scores}).

\noindent \textbf{Topical departure does not imply disruptiveness.}
% Does individual-level exploration also translate into field-level change?
Do papers that move farther from their authors' prior trajectories also redirect subsequent research away from existing research? 
Prior work therefore distinguishes between \emph{disruptive} contributions, which make prior approaches less central to subsequent work, and \emph{consolidating} contributions, which reinforce and extend existing streams of knowledge~\citep{park2023papers,funk2017dynamic}. 
We quantify this distinction using the CD index~\citep{funk2017dynamic}.
For a focal paper $p$, let $n_{\mathrm{focal}}$ denote later papers that cite $p$ but not its references, $n_{\mathrm{both}}$ papers that cite both $p$ and its references, and $n_{\mathrm{prior}}$ papers that cite $p$'s references but not $p$: $\mathrm{CD}_p =
\frac{n_{\mathrm{focal}}-n_{\mathrm{both}}}
{n_{\mathrm{focal}}+n_{\mathrm{both}}+n_{\mathrm{prior}}}.$
Positive values indicate \emph{disruption}, where subsequent work increasingly builds on the focal paper while bypassing its predecessors. Negative values indicate \emph{consolidation}, where the focal paper and its predecessors form coherent themes and continue to be referenced jointly. 
Across strategies, greater topical departure does not correspond to a higher probability of disruption: the share of papers with positive CD falls from $54.0\%$ in the lowest-departure bin to $46.9\%$ in the highest (Figure~\ref{fig:topic_diversity}d). 
However, this relationship differs by research strategy. 
From intermediate to high departure, this share falls from $63.6\%$ to $41.2\%$ for exploiters but rises from $47.6\%$ to $57.7\%$ for cautious explorers. 
The latter increase appears in all three simulated worlds. 
Explorers have the lowest share overall ($39.8\%$), never exceeding $41.3\%$ in any departure bin. 
Changing one's own research direction is therefore not equivalent to changing the direction of the field. Exploration grounded in prior expertise is more likely to produce departures that subsequent research builds on. 
% These differences arise from uptake rather than from how papers are cited. Once cited, $94\%$ to $99\%$ of papers in every departure bin are disruptive, with similar mean CD across strategies (Appendix~\ref{app:disruption}).
\begin{figure*}[t]
    \centering
    \includegraphics[width=0.98\linewidth]{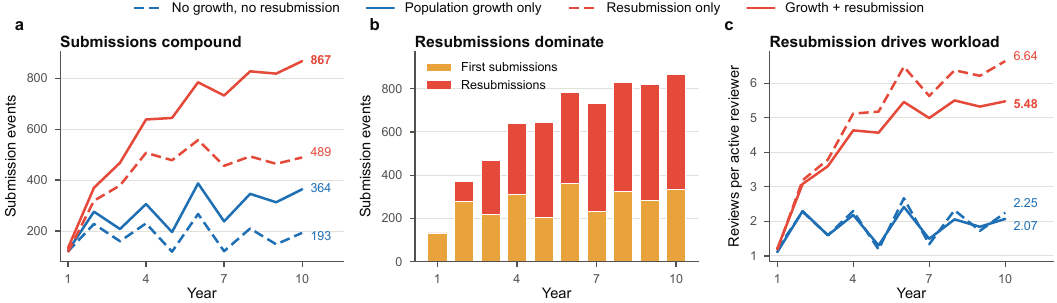}
    \vspace{-.1in}
    \caption{\textbf{Population growth and resubmission compound submission volume but have different effects on reviewer workload.} 
    We cross researcher population growth and resubmission over ten simulated years. \textcolor{figblue}{Blue} / \textcolor{figred}{red} denotes disabling / enabling resubmission, and dashed / solid lines denote without / with  population growth, respectively.
    \textbf{(a)} Annual submission volume. The joint condition produces more submissions than the sum of the increases in either mechanism alone. 
    \textbf{(b)} Submission composition in the joint condition. Resubmissions account for $61\%$ of year-$10$ submissions. 
    \textbf{(c)} Per-reviewer workload. Population growth alone leaves per-reviewer workload close to baseline, whereas conditions with resubmission show substantially higher workload. 
    }
    \label{fig:influx-factorial}
    \vspace{-4mm}
\end{figure*}

% \textbf{Greater selectivity deepens the recycling cascade.}
\noindent \textbf{Resubmission pressure intensifies under greater selectivity.}
We compare matched ten-year simulations that allow rejected papers to be resubmitted: one uses a fixed $30\%$ acceptance rate, and the other follows ACL main-conference acceptance rates from 2016 ($28\%$) to 2025 ($20.3\%$)~\citep{csconfstats}.
The two regimes produce similar numbers of first submissions over the decade ($1{,}706$ vs.\ $1{,}741$), but diverge substantially in total submission growth: $282\%$ under fixed acceptance versus $358\%$ under declining acceptance. 
Rejected papers also accumulate more venue attempts under the declining regime (Figure~\ref{fig:resubmission}b), averaging $3.93$ attempts versus $3.78$ under fixed acceptance.
Mean annual reviews among reviewers assigned at least one paper rise from $1.7$ to $5.5$ under declining acceptance, compared with $1.8$ to $4.6$ under fixed acceptance.
Among papers accepted after resubmission, prior rejections reach ${\sim}3$ by the final year in both regimes (Figure~\ref{fig:resubmission}c).
% These results highlight a dynamic consequence of greater selectivity: when rejected papers are resubmitted, stricter acceptance criteria not only filter more submissions in the current round but also shift review activity into future rounds. This carryover substantially increases reviewer workload even when new manuscript production changes little. 
This comparison shows how greater selectivity can increase review workload without a corresponding increase in new manuscripts: rejected papers shift review activity into future rounds. 
Further results on repeated submission attempts are provided in Appendix~\ref{app:resubmission-extra} (Figure~\ref{fig:resubmission}). 
% We further contrast the fixed 30\% regime with venues that follow the declining main-conference acceptance rates from 2016--2025~\citep{csconfstats}, decreasing from 28\% in the first round to 20.3\% in the tenth (1{,}706 and 1{,}741 first submissions under the fixed and declining regimes). Greater selectivity amplifies resubmission rather than reducing submission pressure: over the decade, submissions grow by 358\% under declining acceptance versus 282\% under fixed acceptance, and cascade depth rises from 3.78 to 3.93 venue attempts while the resubmission rate remains similar (82.1\% vs.\ 84.0\%, Figure~\ref{fig:resubmission}). Repeated resubmission also makes eventual acceptance harder: among papers accepted after rejection, prior rejections rise from about one early in the simulation to roughly three by the final year in both regimes. Reviewer workload grows from 1.8 to 4.6 papers per active reviewer under fixed acceptance and from 1.7 to 5.5 under declining acceptance, and resubmissions score consistently lower than first submissions (mean review score 2.24 vs.\ 2.47).

\noindent \textbf{Revealing author names and institutions slightly increases LLM review scores.}
We evaluate each reviewer--paper pair under three conditions: 1) double-blind review, 2) disclosure of the first author's pseudonymous name and institution, and 3) disclosure of (2) plus a description of the reviewer--author relationship in the collaboration network.
Author-information disclosure raises scores by $0.055$ points on the $1$--$5$ scale.
A separate $120$-paper follow-up, with this contrast prespecified as its primary comparison, finds a $0.077$-point increase ($95\%$ CI $[0.026,0.128]$). 
Additional studies are in Appendix~\ref{app:network-mechanism}.

\subsection{Institutional Mechanisms: Funding Capacity and Allocation}
\label{sec:results-institution}
We examine how institutional resources support scientific production and continued participation. First, we test whether funding capacity keeps pace with higher research output. Then, we ask whether narrowly securing an early grant creates a lasting funding advantage.

\begin{figure*}[t]
    \centering
    \includegraphics[width=0.98\linewidth]{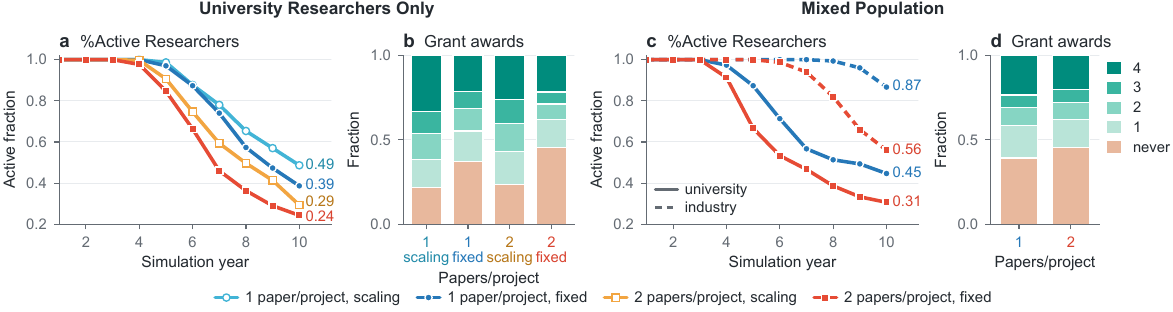}
    \vspace{-2mm}
    \caption{
    We vary submission volume per completed project. 
    \textbf{(a,b)} University-researcher-only simulations compare scaling versus fixed grant capacity.
    \textbf{(c,d)} Mixed-population simulations (university and industry researchers) use fixed grant capacity and a fixed industry compensation pool.
    \textbf{(a,c)} Fraction of the initial researcher population remaining active over years. In (c), solid and dashed lines denote university and industry researchers.
    \textbf{(b,d)} Distribution of cumulative grant awards among the initial university researchers. ``4'' denotes four or more awards and ``never'' denotes no awards.
    }
    \label{fig:scale-main}
    \vspace{-3mm}
\end{figure*}

\noindent \textbf{Higher research output expands publication volume faster than ecosystem capacity.}
AI-assisted research may enable researchers to produce more papers from each completed project, but the resources needed to sustain this additional activity may not expand accordingly. 
Does increasing individual research output necessarily benefit the scientific ecosystem? 
To separate the effect of higher output from differences in research funding modes, we first study university-researcher-only worlds by increasing the submission limit from one to two papers per completed project, under grant capacity that either scales with demand or remains fixed. 
Doubling submission intensity increases accepted papers by $69\%$ under scaling capacity and $63\%$ under fixed capacity, but sharply reduces long-run participation. 
By year ten, the fraction of initial researchers still active falls from $49\%$ to $29\%$ under scaling capacity and from $39\%$ to $24\%$ under fixed capacity (Figure~\ref{fig:scale-main}a). 
These results illustrate a potential risk of increasing research output without adapting the mechanisms that sustain researcher participation. 
Even when grant capacity scales with demand, higher submission intensity produces more accepted papers but leaves fewer researchers active, with little improvement in citation coverage or topic diversity.

Fixed grant capacity primarily reduces access to funding, with smaller changes in inequality among recipients (Figure~\ref{fig:scale-main}b). 
Compared with capacity that scales with demand, the share of researchers who never receive an award rises from $22\%$ to $37\%$ with one submission per project and from $24\%$ to $46\%$ with two submissions.
The Gini coefficient among funded researchers increases only modestly, from $0.36$ to $0.39$ and from $0.40$ to $0.43$, respectively.
Resource scarcity thus manifests mainly in more researchers receiving no funding, rather than greater concentration among recipients. 
The same pattern appears in the mixed university--industry population (Figure~\ref{fig:scale-main}c,d).
Doubling submission intensity increases accepted papers by $77\%$, but reduces year-ten participation from $45\%$ to $31\%$ among university researchers and from $87\%$ to $56\%$ among industry researchers. 
Larger initial resource endowments delay industry attrition but do not prevent it under a fixed compensation pool.
Among university researchers, the share never funded rises from $39\%$ to $45\%$, while inequality among recipients changes modestly.
Citation coverage decreases from $60\%$ to $57\%$, and topic entropy changes little ($0.69$ to $0.71$). 
Overall, publication growth and modest changes in inequality among funding recipients can conceal a broader contraction in participation and funding access. 
% Across both settings, expanding publication output therefore increases scientific production without sustaining a correspondingly larger research community.

% If award-begets-award feedback is absent at the funding margin, does the simulated ecosystem nonetheless stratify on its own?
% In three large unperturbed worlds (1{,}000 institutions and 5{,}000 researchers each, run for a full simulated decade), the funding-resource Gini climbs from near-parity ($\approx0.04$) to $\approx0.36$ by year ten while normalized topic entropy stays high and review standards neither drift nor homogenize (Appendix~\ref{sec:results-ecosystem}, Figure~\ref{fig:ecosystem_dynamics}).
% With the award-to-award channel switched off by construction, the concentration that still emerges must run through accumulated publication and citation records rather than through funding directly rewarding past funding, a distinction that is invisible in aggregate inequality statistics but that the simulator makes explicit.

\subsection{Emergent Ecosystem-Level Dynamics}
\label{sec:results-overall}

The study in Appendix~\ref{sec:results-matthew} finds no detectable subsequent-funding advantage.
This raises a broader question: \emph{can persistent inequality nevertheless emerge endogenously from the coupled dynamics of scientific production, citation, funding, and researcher attrition?}
To examine this question, we run three large confirmatory worlds, each with $1{,}000$ institutions and $5{,}000$ researchers, for a full simulated decade without experimental intervention. 
We track the evolution of resource inequality, topic diversity, active population size, and peer-review dynamics (Appendix~\ref{sec:results-ecosystem}; Figure~\ref{fig:ecosystem_dynamics}). 

\noindent \textbf{Substantial stratification emerges over time.}
Funding-resource inequality increases steadily over time: the Gini coefficient rises from near parity at initialization ($\approx0.04$) to $\approx0.36$ by year ten (Figure~\ref{fig:ecosystem_dynamics}a). 
Citation inequality remains consistently high throughout the simulation ($\approx0.56$ to $\approx0.54$). 
Importantly, this concentration of resources does not coincide with a collapse in research diversity. 
New research-direction choices remain broadly distributed across topics, with topic entropy remaining high at ${\sim}0.81$ to ${\sim}0.85$. By the end of the simulation, all $53$ available research directions attract active researchers (Figure~\ref{fig:ecosystem_dynamics}b). This reveals a decoupling between resource inequality and intellectual diversity: resources become increasingly concentrated even as the research landscape remains broadly diversified.
At the same time, inequality is accompanied by substantial researcher attrition. 
The active population declines from $5{,}000$ researchers to approximately $2{,}530$ by year ten as resource-depleted researchers exit the system (Figure~\ref{fig:ecosystem_dynamics}c). 
Peer-review dynamics remain comparatively stable. 
Across $455$K individual reviews, the mean score remains between $2.29$ and $2.40$, with $62$--$72\%$ of scores falling between $2.0$ and $2.5$. 
The standard deviation across individual reviews increases slightly from $0.50$ to $0.55$, while mean disagreement among reviewers of the same paper increases from $0.29$ to $0.35$ (Figure~\ref{fig:ecosystem_dynamics}d).

\vspace{-1mm}
\section{Related Work}
\vspace{-1mm}
\noindent \textbf{Large language model agents and agent-based models of science.}
Recent systems use LLM agents for idea generation~\citep{lu2024ai,tang2026ai}, experimentation~\citep{yamada2025ai}, manuscript preparation~\citep{weng2025cycleresearcher}, and peer review~\citep{jin2024agentreview,yang2026paper}.
In parallel, a longer tradition of agent-based models studies how individual incentives and institutional rules generate collective outcomes~\citep{vsevselja2022agent}, including publication and resubmission~\citep{kovanis2016complex}, resource-constrained publishing and reviewing~\citep{bianchi2022can},  long-term effects of publication incentives~\citep{smaldino2016natural},
funding and cumulative advantage~\citep{merton1968matthew,bol2018matthew}, and risks from adversarial participants~\citep{jiang2026badscientist}. 
\method couples research choice, collaboration, publication, peer review, resubmission, citation, funding, resources, and career survival within a persistent, literature-grounded LLM ecosystem, enabling controlled studies of how local mechanisms propagate into longer-run outcomes.

\noindent \textbf{Large language model-based social simulation.}
LLM-based simulations study complex systems that are difficult to manipulate directly~\citep{anthis2025position,wang2026mascot,wang2025companioncast}. 
These provide controlled tests of how differences emerge and accumulate over time. This is particularly important as LLM generations increasingly influence and mutually reinforce human writing and communication~\citep{kobak2025delving,yakura2024empirical,zhu2024divergent,jin2026unisd}.
Closer to our setting, recent work moves toward research-community simulation: AgentRXiv~\citep{schmidgall2025agentrxiv} lets agents build on prior work across repeated interactions, while ResearchTown~\citep{yu2024researchtown} models researchers and papers as an agent--data graph for reading, writing, and reviewing.
\vspace{-1mm}
\section{Conclusion}
\vspace{-1mm}
We presented \method, a holistic, LLM-based framework for simulating academic research as an evolving ecosystem. 
By integrating research-direction choice, collaboration, submission, peer review, resubmission, citation, funding, memory, and attrition into a single literature-grounded loop, \method helps policy-makers understand how local decisions accumulate into long-term scientific dynamics. 
It provides a testbed for asking not only how scientific agents behave, but how the rules of academia shape what research survives, spreads, and ultimately succeeds.

\FloatBarrier
\bibliographystyle{unsrtnat}
% Keep long bibliography URLs from stretching the surrounding word spacing.
\begingroup
\raggedright
\bibliography{references}
\endgroup

\clearpage
\appendix
\raggedbottom

\FloatBarrier
\section{Human Factors: Exploration Strategies and Careers}
\label{app:human-factors}

\FloatBarrier
\subsection{Cautious exploration balances novelty with career reward}
\label{app:strategy-careers}

\noindent \textbf{Reported worlds.}
The individual-strategy study uses five calibration worlds % with $100$ institutions 
and three confirmatory worlds with $1{,}000$ institutions over ten years. Each institution has five researchers: one explorer, three exploiters, and one cautious explorer, initialized with equal funding and reputation. Figure~\ref{fig:exploration_strategies}a,b uses the calibration worlds, and c,d uses the confirmatory worlds.

\noindent \textbf{Metrics.}
For research strategy, we measure exploration distance (ED), direction-switching, acceptance rate, funding, and citation outcomes. Figure~\ref{fig:exploration_strategies}a reports the probability of producing at least one paper in its cohort's top citation decile. Panel c reports whether a researcher's aggregated citation total reaches the within-world $90$th percentile (ties included).

\noindent \textbf{Strategy manipulation check.}
\label{app:strategy-check}
Figure~\ref{fig:exploration_strategies}a shows the calibration-world manipulation check (mean ED $0.82$, $0.67$, and $0.43$ for explorers, cautious explorers, and exploiters).
The confirmatory worlds show the same ordering: explorers move farthest from their preceding papers (mean ED $0.83$), cautious explorers occupy the middle ($0.69$), and exploiters stay closest to their preceding papers ($0.49$).
% Thus the outcome comparisons below reflect different search regimes, rather than noisy variants of the same behavior.

\FloatBarrier
\subsection{Cautious exploration sustains long-run scientific diversity}
\label{app:strategy-diversity}

\noindent \textbf{Reported worlds.}
The study that varies the proportions of the three strategies uses ten matched triples of $100$-institution worlds with explorer-heavy ($3/1/1$), exploiter-heavy ($1/4/0$), or cautious-heavy ($1/1/3$) compositions.

\noindent \textbf{Metrics.}
For worlds that differ in the proportions of the three strategies, we measure the area under the ten-year normalized topic-entropy curve (defined in Eq.~\ref{eq:topic-entropy}), the number of occupied research directions, and total submissions. % Entropy and occupied-direction counts summarize new direction choices in each year, with each choice weighted equally regardless of project duration.

\FloatBarrier
\section{Research Outcomes: Impact, Resubmission, and Peer Review}
\label{app:research-outcomes}

\FloatBarrier
\subsection{Disruption Depends on Uptake}
\label{app:disruption}

\begin{figure}[!htbp]
\centering
\includegraphics[width=0.8\linewidth]{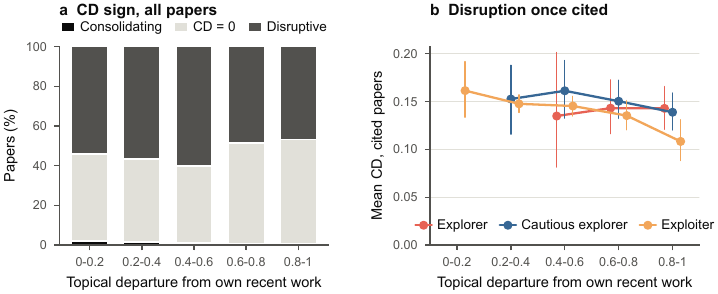}
\caption{\textbf{Disruption tracks whether a paper is cited at all.}
\textbf{(a)} CD of the $8{,}507$ papers at age three, pooled over three confirmatory worlds. Almost all papers with CD $=0$ ($98.5\%$) have not yet been cited, and consolidating papers (CD $<0$) never exceed $2.0\%$ of a departure bin.
\textbf{(b)} Mean CD among cited papers with bootstrap $95\%$ confidence intervals. Once cited, papers from all three strategies are similarly disruptive.}
\label{fig:disruption_uptake}
\end{figure}

Figure~\ref{fig:disruption_uptake} supports the disruption analysis in \S\ref{sec:results-research}.
All three citation counts in $\mathrm{CD}_p$ use the same observation window, from the focal paper's publication year through age three. We report CD only for papers with references and at least three papers in the denominator.

\FloatBarrier
\subsection{Submission growth and reviewer workload under population change and resubmission}
\label{sec:results-resubmission}
% Rapid submission growth in areas such as AI and machine learning has strained peer-review systems, increasing reviewer workload and threatening high-quality evaluation. This growth is often attributed to 1) an expanding research population and 2) rejected manuscripts returning to review~\citep{yang2026paper}.
Rapid growth in conference submissions has increased pressure on peer-review systems, especially in AI and machine learning.
Two mechanisms are commonly implicated: growth in the researcher population and the resubmission of rejected manuscripts~\citep{yang2026paper}.
Although both can increase submission volume, they imply different dynamics: population growth introduces more new manuscripts and more researchers, whereas resubmission repeatedly returns existing manuscripts to review.
% What drives submission inflation: more researchers, rejection recycling, or their interaction? And which mechanism most increases reviewer burden?
We therefore ask %how population growth and resubmission relate to total submission volume and per-reviewer workload, and whether their joint pattern departs from the sum of their individual contributions.
whether review pressure arises primarily from growth in new submissions or from repeated resubmission.

% To separate these effects,
We run a $2{\times}2$ counterfactual experiment crossing researcher population growth with resubmission, while holding the initial population, venues, literature corpus, and acceptance rate fixed.
Under population growth, institutions add new researchers each year equal to $10\%$ of the initial population. Under resubmission, rejected papers may return to review at a fixed resource cost (Appendix~\ref{sec:eval-setup}).
Both mechanisms increase submission volume, but their combination produces more submissions than the sum of their separate effects (Figure~\ref{fig:influx-factorial}a).
By year $10$, compared with $193$ in the baseline, annual submissions reach $364$ under population growth alone and $489$ under resubmission alone. When both mechanisms are enabled, submissions rise further to $867$.
Resubmissions account for $61\%$ of year-$10$ submissions in this joint condition (Figure~\ref{fig:influx-factorial}b).
% Resubmissions also make up 61\% of all year-10 submissions in the joint condition (Figure~\ref{fig:influx-factorial}b), showing how rejected manuscripts can remain in the submission pool across multiple years.
% The joint world produces 207 more than this sum, so the interaction between the two mechanisms accounts for 31\% of the total increase over the baseline.

Total submission volume, however, does not directly determine reviewer workload.
Despite substantially more submissions, population growth alone leaves reviews per active reviewer close to baseline ($2.07$ vs.\ $2.25$ at year $10$), as the larger researcher population also distributes the review workload across more reviewers (Figure~\ref{fig:influx-factorial}c).
With resubmission, workload instead rises to $6.64$ reviews per active reviewer without population growth and $5.48$ when both mechanisms are enabled. These ratios divide all annual reviews by the year-end active reviewer pool.
Thus, similar increases in submission volume can have very different reviewer-level consequences depending on whether they come from new researchers or manuscripts repeatedly returning after rejection.
% Additional results are provided in Appendix~\ref{app:resubmission-extra}.

To quantify this persistence, we define the resubmission multiplier $M_t$ as the ratio of total submissions to first submissions in year $t$.
By year $10$, $M_t$ reaches $2.78$ under resubmission alone and $2.59$ when combined with population growth (Figure~\ref{fig:resubmission}a), i.e., each first submission is accompanied by roughly $1.6$--$1.8$ additional resubmissions at the system level.
Among papers rejected at least once, the average number of venue attempts reaches $3.84$ under resubmission alone and $3.58$ in the joint condition, while $84.2\%$ and $81.8\%$ of rejected papers, respectively, are resubmitted rather than abandoned.
Under the modeled resubmission cost, these repeated attempts are also accompanied by greater researcher attrition: without population growth, $221$ of $302$ researchers remain active at year $10$ under resubmission versus $257$ without it; with population growth, $475$ versus $527$ of the final $600$ researchers remain active.
% Under the modeled resubmission cost, these repeated attempts are also accompanied by greater researcher attrition: without population growth, 221 of 302 researchers remain active at year 10 under resubmission versus 257 without it; with population growth, 475 versus 527 of the final 600 researchers remain active.

\noindent \textbf{Reported worlds.}
The resubmission study uses four matched $10$-year worlds crossing researcher population growth and rejection recycling, plus a separate pair comparing fixed and declining acceptance rates. Under population growth, institutions alternately add junior researchers, producing annual entry equal to $10\%$ of the initial population. Entrants receive standard initial resources and expertise sampled from the founding distribution.

\noindent \textbf{Metrics.}
For resubmission, we measure submission volume, submission composition (first submissions vs.\ resubmissions), the resubmission multiplier, cascade depth, time to publication, reviewer workload (reviews per researcher active at year end), venue redistribution, and review-score differences between first submissions and resubmissions. Each submission receives three reviews. We track the frequency and persistence of resubmission, their interaction with population growth, and the changing active reviewer population.

\FloatBarrier
\subsection{Additional Resubmission-Cascade Results}
\label{app:resubmission-extra}

\begin{figure}[!htbp]
\centering
\includegraphics[width=\linewidth]{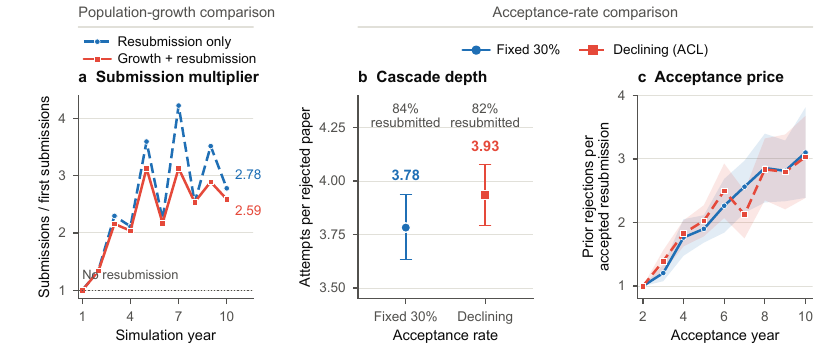}
\caption{\textbf{Recycling multiplies submissions, and resubmission cascades deepen under stricter acceptance.}
\textbf{(a)} Annual multiplier $M_t$: all submission events divided by first submissions. Year-ten values are $2.78$ without population growth and $2.59$ with growth; the dotted line marks $M_t=1$ without resubmission.
\textbf{(b, c)} A separate matched pair of ten-year worlds compares fixed $30\%$ acceptance with declining ACL-calibrated rates.
\textbf{(b)} Mean venue attempts per paper rejected at least once, before acceptance or abandonment. Percentages indicate the fraction resubmitted; bars show paper-level bootstrap $95\%$ confidence intervals.
\textbf{(c)} Mean prior rejections among resubmitted papers accepted in each year; shading shows the mean $\pm 1.96$ standard errors. Each condition has one world; intervals do not estimate variability across worlds.}
\label{fig:resubmission}
\end{figure}

Figure~\ref{fig:resubmission} supports the resubmission analysis in \S\ref{sec:results-resubmission}.

\FloatBarrier
\subsection{No-influx resubmission comparison}
\label{app:large-resubmission}

This separate comparison uses one complete ten-year world per condition with the same $1{,}202$ founders: five researchers at each of $240$ universities and two industry researchers. Condition A disables resubmission and C enables it. Neither admits new researchers. Both use Qwen3-32B, three reviews per submission, and panelized sequential funding selection from the remaining application IDs. Panels contain at most $25$ applications, with acceptance rates of $0.23$/$0.10$ for NSF/DARPA and quota $\max(1,\lfloor n_{\mathrm{panel}}r\rfloor)$. 

\begin{figure}[H]
\centering
\includegraphics[width=\linewidth]{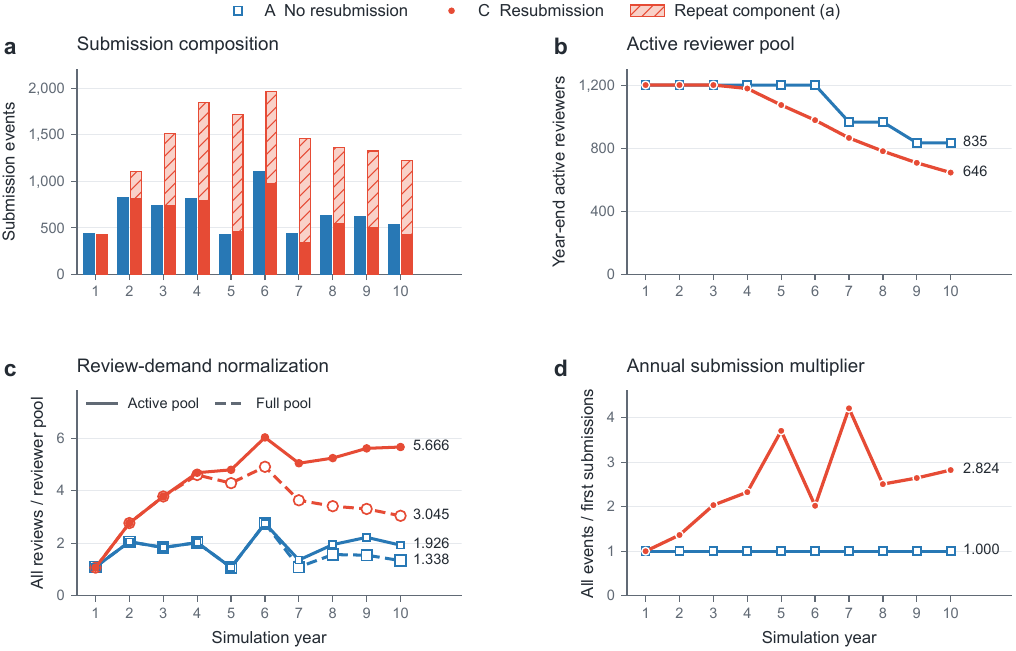}
\caption{\textbf{Annual submission and review demand in the A/C comparison.}
\textbf{(a)} Paired annual bars (A left, C right) separate first submissions (solid) from repeats (hatched). Bar heights count all submission events.
\textbf{(b)} Year-end reviewer pool $N_t$, counting researchers satisfying both \texttt{is\_active} and \texttt{can\_review}.
\textbf{(c)} All individual reviews $R_t$ divided by the active pool (solid lines) or the full $1{,}202$-person pool (dashed lines), on a shared axis. The numerator includes reviews by researchers inactive at year end, so active-pool ratios are not mean assigned workload among survivors.
\textbf{(d)} Annual total/first-submission multiplier $M_t$.
Each submission event receives three reviews, so review demand is mechanically linked to the submission totals in (a).
Points are saved annual observations from one ten-year world per condition (no influx). Lines connect observations without smoothing. No across-world uncertainty interval is available from this pair.}
\label{fig:large-resubmission}
\end{figure}

We measure review demand relative to the active research community as $R_t/N_t$, where $R_t$ is the total number of reviews in year $t$ and $N_t$ is the number of researchers active at year end.
We also normalize review counts by the initial population of $1{,}202$ researchers to distinguish changes in review volume from the effects of attrition.
The annual resubmission multiplier $M_t$ is the ratio of total submissions to first submissions within the same year.
Figure~\ref{fig:large-resubmission} traces these measures and submission composition over ten years.

In year ten, C records $3{,}660$ individual reviews versus A's $1{,}608$. Division by the year-end active reviewer pool gives $5.666$ versus $1.926$, while division by the full $1{,}202$-person pool gives $3.045$ versus $1.338$. C has fewer first submissions ($432$ versus $536$), plus $788$ repeat submissions. C has higher review volume and both normalized measures in years $2$--$10$, but fewer reviews in year one ($1{,}287$ versus $1{,}320$), when neither world has any resubmission. Shared founders do not imply identical realized pre-resubmission trajectories. C's review volume and active-pool normalization peak in year six ($5{,}904$ and $6.031$). Neither grows monotonically.

Over ten years, C has $6{,}040$ first submissions versus A's $6{,}566$, plus $7{,}912$ repeats, yielding $13{,}952$ versus $6{,}566$ submission events and $41{,}856$ versus $19{,}698$ reviews. Each event receives three reviews, so these volumes are mechanically linked, not independent evidence of greater research production or quality. The smaller year-ten active pool in C ($646$ versus $835$) amplifies the primary ratio, while the fixed-pool difference remains positive. These counts do not establish that reviewing caused inactivity. This pair supports a descriptive review-demand contrast, not an influx interaction, an isolated funding mechanism, or independent world-level uncertainty estimates.

\FloatBarrier
\subsection{Details of the Author-Information Experiment}
\label{app:network-mechanism}

\noindent \textbf{Reported worlds.}
The network-proximity study generates valid collaboration networks and replays matched peer-review decisions under controlled information conditions.

\noindent \textbf{Review conditions.}
We separate the effect of revealing author information from the additional effect of explicitly disclosing the reviewer--author relationship in the collaboration network.
Each reviewer evaluates the same paper under three conditions:
\begin{enumerate}[leftmargin=*,itemsep=0pt,topsep=3pt]
    \item \textbf{Condition 1: Blind.} No author name, institution, or network relationship is shown.
    \item \textbf{Condition 2: Author information.} The first author's name, represented by a stable pseudonymous identifier, and institution are shown, but the network relationship is not disclosed.
    \item \textbf{Condition 3: Author information plus relationship.} The same author information is shown alongside a neutral statement of the actual network relationship, such as sharing an intermediary collaborator or belonging to disconnected components.
\end{enumerate}
Across conditions, we hold fixed the paper's title, abstract, and topic labels, the reviewer's expertise, the scoring instructions, and the $1$--$5$ scale.

\noindent \textbf{Papers and reviewer matching.}
For each paper, we select three reviewers with similar expertise but different relationships to the first author: a distance-$2$ reviewer shares a collaborator with the author (distance $2$); a distance-$3$ reviewer is connected through two intermediary researchers; and a stranger is connected to the author through a shortest-path with distance $\geq4$ or in a disconnected component.
We exclude direct collaborators, coauthors, researchers affiliated with any author's institution, and researchers with recorded conflicts of interest.
Within each triplet, paper--reviewer cosine similarities differ by at most $0.10$, and each reviewer is assigned at most ten papers.

\noindent \textbf{Collaboration network.}
Network distances are computed using only collaborations formed before each paper's review year.
The source simulations allow both new and repeated partnerships: in each agent-year, collaborator search targets new partners with probability $0.70$ and existing partners otherwise.
New-partner scores combine topic similarity, recent-publication overlap, log degree, and shared collaborators, with respective weights of $0.40$, $0.30$, $0.20$, and $0.10$.
The resulting network supplies the relationships used for reviewer matching and disclosure.

\noindent \textbf{Score comparisons.}
For network-proximity bias, the primary outcome is the paper-level matched score shift caused by disclosing the reviewer's true network relationship to the author.
We compute two within-reviewer, within-paper differences.
The \emph{author-information effect} is the score under author-information disclosure (Condition 2) minus the score under blind review (Condition 1).
The \emph{additional relationship effect} is the score under author-information-plus-relationship disclosure (Condition 3) minus the score under author-information disclosure alone (Condition 2).
For each effect, we first average the three reviewers' score changes within each paper, then average across papers.
We report changes on the $1$--$5$ scale, with $95\%$ confidence intervals from $10{,}000$ bootstrap resamples of whole papers.

We also subtract the stranger reviewer's relationship-induced score change from that of the reviewer who shares a collaborator with the author.
This tests whether disclosing a closer relationship produces a larger score change.

\noindent \textbf{Results and interpretation.}
\label{sec:results-network-bias}
Revealing author information alone increased the mean score by $0.055$ points ($95\%$ CI $[0.014,0.097]$).
% The 120-paper follow-up tested this finding on papers from a new simulation world, making author-information disclosure its main comparison.
% It found a $0.077$-point increase (95\% CI $[0.026,0.128]$, paired permutation $p=0.004$).
Once author information was visible, adding a description of the relationship had little average effect, with a mean score change of $0.000$ points ($95\%$ CI $[-0.038,0.037]$).
Reviewers who shared a collaborator with the author did not show a detectably larger response to relationship disclosure than stranger reviewers.

\FloatBarrier
\section{Institutional Mechanisms: Funding Capacity and Allocation}
\label{app:institutional-mechanisms}

\FloatBarrier
\subsection{Higher research output expands publication volume faster than ecosystem capacity}
\label{app:output-scaling}

\noindent \textbf{Reported worlds.}
The output-scaling study uses four university-only worlds with $300$ researchers each, crossing one or two submissions per completed project with scaling or fixed grant capacity. Two mixed-population worlds, each with $150$ university and $150$ industry researchers, compare the submission limits under fixed capacity. All six worlds run for ten years. Scaling capacity applies each funding program's configured award rate to its application volume; fixed capacity reserves annual award slots equal to $20\%$ of the initial university population, giving $60$ slots in university-only worlds and $30$ in mixed worlds. The mixed worlds also hold the industry compensation pool fixed.

\noindent \textbf{Output intervention.}
The intervention varies the submission limit per completed project while retaining the same cost settings, project-duration rules, and manuscript-selection procedure. Table~\ref{tab:implementation-settings} reports the resource parameters.

\noindent \textbf{Metrics.}
For output scaling, active fractions use the initial population as the denominator. Grant coverage and award-count inequality include all initial university researchers, including those inactive by year ten; inequality among recipients conditions on receiving at least one award. Three-year citation coverage is the share of accepted papers receiving at least one citation from a paper with a disjoint author set within three years, restricted to papers accepted by year seven. Topic entropy is rarefied to a common accepted-paper sample size within each population comparison. Citation opportunities also depend on submission volume and the size of the accepted-paper pool, so these citation measures are descriptive rather than direct estimates of scientific quality.

\FloatBarrier
\subsection{Narrow funding wins show no detectable increase in later funding}
\label{sec:results-matthew}
Prior work compares applicants near grant cutoffs to study how early funding success or failure shapes later outcomes~\citep{bol2018matthew,wang2019early}.
We use \method to trace these effects over three years following researchers' first application year.
We call applicants ranked just above a panel's funding cutoff \emph{narrow winners} and those just below it \emph{narrow losers}.
Their adjacent panel ranks provide a local comparison of subsequent funding, publications, citations, and research choices.

\begin{figure*}[!htbp]
\centering
\includegraphics[width=0.98\linewidth]{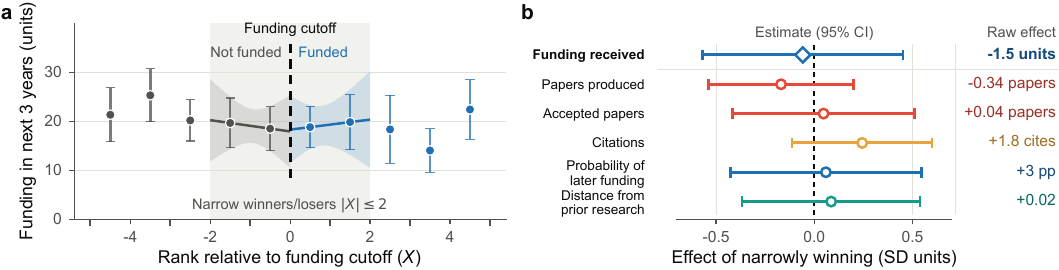}
\vspace{-1.5mm}
\caption{\textbf{Narrow funding wins show no detectable advantage over the next three years.}
\textbf{(a)} Applicants just above the funding cutoff (\emph{narrow winners}) receive similar subsequent funding to those just below it (\emph{narrow losers}), with no clear jump at the cutoff.
Points show mean funding at each rank. Lines are descriptive local fits, and shading marks the near-cutoff comparison window ($N=444$ applications).
\textbf{(b)} Estimated effects on later funding, publications, citations, award probability, and exploration all have confidence intervals spanning zero.
Colors group outcome families, and the diamond marks the primary funding outcome.
Effects are scaled by the narrow-loser standard deviation, with raw-unit estimates at right (pp: percentage points).
Error bars and bands show $95\%$ confidence intervals.}
\label{fig:matthew_rd}
\vspace{-3mm}
\end{figure*}

Narrow winners show no detectable increase in subsequent funding (Figure~\ref{fig:matthew_rd}a): the estimated effect is $-1.50$ resource units ($95\%$ CI $[-14.5,+11.5]$).

Publication output, citations, future award probability, and exploration likewise show no detectable differences (Figure~\ref{fig:matthew_rd}b).
All near-cutoff applicants reapply, and none exits the simulation during follow-up.
The comparison follows applicants who remain in funding competition throughout the three-year window.

\noindent \textbf{Reported worlds.}
The funding-cutoff study uses three $500$-agent, six-year worlds and records within-panel ranks for regression-discontinuity analysis.

\noindent \textbf{Estimation and diagnostics.}
The running variable is $x=k-\mathrm{rank}+0.5$, where $k$ is the panel's award count, so winners have $x>0$. We fit a local-linear model with separate slopes on either side, triangular weights, and competition-panel fixed effects. The primary bandwidth $h=2$ is selected by leave-one-competition-out cross-validation on pre-application publication count. The sample contains $444$ applications from $265$ applicants in $183$ competitions. Confidence intervals use two-way cluster-robust standard errors by applicant and competition, with finite-sample corrections. Bandwidth checks at $h\in\{2,3,4,6,8\}$ yield funding estimates from $-1.50$ to $+0.47$, all with intervals spanning zero. Balance checks detect no cutoff differences in prior paper or acceptance counts, concurrent applications, or strategy indicators.

\noindent \textbf{Metrics.}
For funding RD, the primary outcome is cumulative subsequent funding over the next three simulated years; secondary outcomes include publication, acceptance, citation, reapplication, attrition, and exploration behavior.

\FloatBarrier
\section{Emergent Ecosystem-Level Dynamics}
\label{app:ecosystem-dynamics}

\FloatBarrier
\subsection{Resource Inequality and Researcher Attrition}
\label{sec:results-ecosystem}

\begin{figure*}[!htbp]
\centering
\includegraphics[width=0.98\linewidth]{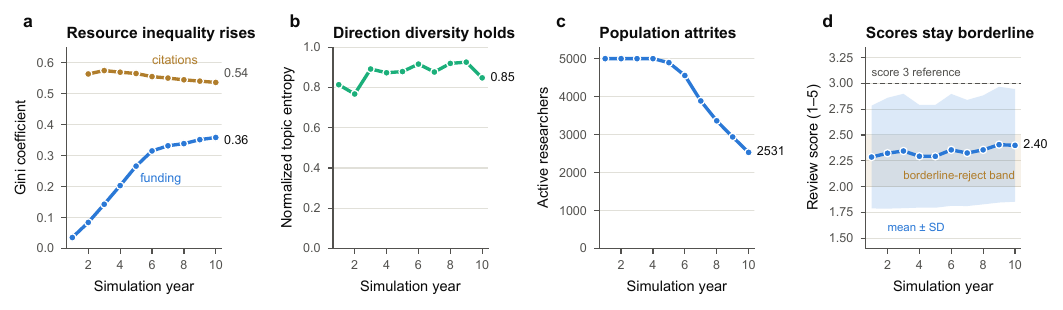}
\caption{\textbf{Resource inequality compounds while topic diversity holds.} Pooled over three confirmatory worlds ($1{,}000$ institutions, $5{,}000$ researchers, ten years); lines show means and bands span the minimum--maximum across worlds. \textbf{(a)} Funding-resource Gini (rising) with the citation Gini overlaid. \textbf{(b)} Normalized topic entropy, where $1$ denotes research directions being occupied evenly. \textbf{(c)} Number of still-active researchers. \textbf{(d)} Mean individual review score with a $\pm1$~SD envelope. Scores remain between $2.29$ and $2.40$ on the $1$--$5$ scale (median $2.0$ every year), while the envelope widens slightly. The dashed line at $3$ is a score reference. Acceptance follows the ranking-and-quota rule in \S\ref{sec:method}.}
\label{fig:ecosystem_dynamics}
\vspace{-3mm}
\end{figure*}

Figure~\ref{fig:ecosystem_dynamics} supports the ecosystem-level analysis in \S\ref{sec:results-overall}.

\FloatBarrier
\subsection{Self-citation does not explain citation inequality}
\label{app:self-citation}
\begin{figure*}[!htbp]
\centering
\includegraphics[width=0.98\linewidth]{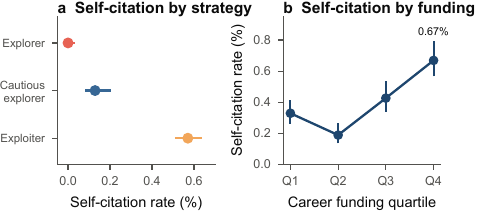}
\caption{
\textbf{Self-citation across strategies.} Across three confirmatory worlds ($75{,}661$ citation edges), exploiters self-cite more frequently than their peers.
Removing all self-citation edges changes the citation Gini by at most $0.0005$ in any world.
% Error bars show Wilson 95\% confidence intervals.
}
\label{fig:self_citation}
\vspace{-3mm}
\end{figure*}

Because agents may cite their own prior work, we test whether self-citation contributes materially to the citation inequality observed above.
Across the three large confirmatory worlds spanning $1{,}000$ institutions, $5{,}000$ researchers, and ten years, only $308$ of $75{,}661$ citation edges ($0.41\%$) are self-citations (Figure~\ref{fig:self_citation}).
Meanwhile, self-citation differs by research strategy.
Exploiters self-cite in $0.57\%$ of outgoing citation edges, compared with $0.13\%$ for cautious explorers and around $0.0\%$ among explorers ($0$ of $10{,}861$ edges).
This ordering holds in all three worlds.
The pattern is consistent with topical continuity: exploiters remain close to their prior work, whereas explorers move farther away and therefore have fewer relevant prior papers of their own to cite.
Despite these differences, self-citation is too rare to explain the ecosystem's citation concentration, indicating that citation inequality in \method{} arises from broader citation-allocation dynamics.

\noindent \textbf{Citation Inequality Persists After Removing Self-Citations}

Figure~\ref{fig:self_citation} supports the self-citation analysis in \S\ref{sec:results-overall}.

\clearpage
\FloatBarrier
\section{Notation}
\label{app:notation}
Table~\ref{tab:notation} lists the symbols used in the paper, grouped by where they are introduced.

\begin{table}[H]
\centering
\caption{Notation used in the paper.}
\label{tab:notation}
\small
\begin{tabular}{@{}lp{0.6\linewidth}@{}}
\toprule
Symbol & Meaning \\
\midrule
\multicolumn{2}{@{}l}{\emph{Researchers and budgets (\S\ref{sec:method})}} \\
$i$, $j$ & Researcher indices ($j$ also denotes a candidate collaborator or a reviewer) \\
$t$ & Simulated year \\
$s_i^{t}=(E_i,\rho_i^{t},b_i^{t},M_i^{t},C_i^{t})$ & State of researcher $i$ in year $t$: expertise topics, reputation, available budget, memory, and conflict-of-interest set \\
$f_i^{t}$ & Funding received by researcher $i$ in year $t$ \\
$\mathcal{A}_i^{t}$, $\mathrm{cost}(a)$ & Actions taken by researcher $i$ in year $t$ and the cost of action $a$ \\
\midrule
\multicolumn{2}{@{}l}{\emph{Venues, directions, and collaboration (\S\ref{sec:method})}} \\
$c$, $\alpha_c$, $S_c$ & A conference, its acceptance rate, and its submissions in a year \\
$d\in\mathcal{D}$ & A research direction from the fixed set $\mathcal{D}$ ($|\mathcal{D}|=53$) \\
$T_d$, $\kappa$ & Duration of direction $d$ in years and its budget cost per active year \\
$\psi(i,j)$, $\phi_m(i,j)$, $w_m$ & Collaboration score of the pair $(i,j)$, its $m$-th feature, and the feature's weight \\
$p$, $R_p$ & A paper and its reviewer panel \\
$u_{jp}$, $\bar u_p$ & Recommendation of reviewer $j$ on paper $p$ and the panel mean used for the decision \\
$n$, $n_{\mathrm{panel}}$, $r$, $k$ & Panel application count ($n$ or $n_{\mathrm{panel}}$), program funding rate, and award count under rate-based funding, $k=\max(1,\lfloor rn\rfloor)$ for nonempty panels \\
$e_i$, $\tilde e_i$, $\lambda$ & Researcher $i$'s normalized panel-rank score $e_i\in[0,1]$, adjusted score $\tilde e_i=e_i-\lambda q_i$, and penalty strength $\lambda\geq0$ \\
$q_i$ & Percentile of researcher $i$'s mean paper distance among that year's applicants with scored papers, using all recorded papers. Applicants without scored papers receive $q_i=0$ \\
\midrule
\multicolumn{2}{@{}l}{\emph{Evaluation (\S\ref{sec:evaluation})}} \\
$\mathbf{p}$, $\mathbf{a}_i^{t}$ & Paper embedding and researcher $i$'s mean paper embedding from years $t-4$ through $t-1$, with initial expertise as the fallback \\
$\nu_{p,i}^{t}$ & Paper distance $\nu_{p,i}^{t}=1-\cos(\mathbf{p},\mathbf{a}_i^{t})\in[0,2]$ (\S\ref{sec:results-human}) \\
$\mathrm{ED}$ & Mean annual distance to the author's most recent paper from an earlier year, excluding years without a comparison (\S\ref{sec:results-human}) \\
$\mathrm{CD}_p$; $n_{\mathrm{focal}}$, $n_{\mathrm{both}}$, $n_{\mathrm{prior}}$ & CD index and its three citation counts for focal paper $p$, observed at age three: focal-only, focal-and-reference, and reference-only citing papers (\S\ref{sec:results-research}) \\
$H_t$, $p_{t,d}$ & Normalized topic entropy in year $t$ and the share of that year's direction choices selecting $d$, with $0\log0=0$ (Eq.~\ref{eq:topic-entropy}) \\
$\mathrm{AUC}_H$, $T$ & Trapezoidal area under $H_t$ from year $1$ to $T$, bounded by $0$ and $T-1$ (Eq.~\ref{eq:topic-entropy}) \\
$M_t$ & Resubmission multiplier: total submissions divided by first submissions in year $t$ (\S\ref{sec:results-resubmission}) \\
$R_t$, $N_t$ & All individual review events in year $t$ and the year-end active reviewer-capable population, used in $R_t/N_t$ (Appendix~\ref{app:large-resubmission}) \\
$N$ & Number of observations in an analysis sample \\
\bottomrule
\end{tabular}
\end{table}

\FloatBarrier
\section{Experimental Details}
\label{sec:eval-setup}

\FloatBarrier
\subsection{Data}
To ground the simulation in realistic scientific literature, we use \dataset~\citep{jin2024scievo} as our literature corpus.
\dataset is a longitudinal scientometric dataset containing $2.1$ million arXiv papers published from 1991 to 2025, spanning eight broad subject groups and $156$ categories.
It provides paper titles, abstracts, publication timestamps, subject categories, and LLM-extracted keywords, together with citation relations, author information, and publication venues retrieved from Semantic Scholar.
For our experiments, we select computer science papers within our 2016--2025 simulation window and organize them by publication year.

\FloatBarrier
\subsection{Model and literature grounding}
All simulations use Qwen3-32B~\citep{yang2025qwen3} with thinking mode. Our experiments used NVIDIA A100 GPUs with $80$\,GB of memory. 
Submissions are based on retrieval grounded in \dataset.
Given an agent's selected research direction and submission cycle,
the agent retrieves contemporaneous arXiv papers and selects a topically compatible paper and venue.
Simulated years correspond to 2016--2025, with year-based filtering of the retrieved corpus determining which papers are available for submission and citation.
Unless otherwise specified, conferences use a fixed $30\%$ acceptance rate to ensure comparable selectivity across counterfactual worlds.
The resubmission-cascade study (\S\ref{sec:results-resubmission}) uses this regime, while Appendix~\ref{app:resubmission-extra} additionally evaluates a schedule based on observed yearly acceptance rates.

All families use the same set of research directions $\mathcal{D}$.

\FloatBarrier
\subsection{Implementation settings}
\label{app:implementation-settings}
Table~\ref{tab:implementation-settings} records the resource settings used in the reported studies. The capacity treatments, population compositions, and numbers of worlds are specified with each experiment.

\begin{table}[!htbp]
\centering
\caption{Resource and funding settings. Budgets and costs are in abstract resource units.}
\label{tab:implementation-settings}
\small
\begin{tabular}{@{}p{0.36\linewidth}p{0.59\linewidth}@{}}
\toprule
Setting & Value and scope \\
\midrule
Uniform initial budget & $100$ per researcher in the strategy, composition, funding-cutoff, and university-only output studies \\
Default initial endowments & University: $80$ or $160$. Industry: $160$ or $320$, depending on institution tier \\
Research cost & $10$ per active project-year \\
Project duration & $1$--$3$ years, determined by the selected direction \\
Resubmission fee & $5$ per attempt \\
Grant amount and rate & $20$ per award. Foundation-oriented programs fund $23\%$ and mission-oriented programs $10\%$ of panel applicants under rate-based allocation, subject to the quota rule in \S\ref{sec:method} \\
Funding-conservatism penalty & Effective $\lambda=0$ in the reported experiments \\
\bottomrule
\end{tabular}
\end{table}

\noindent \textbf{Retrieval and semantic distance.}
Retrieval uses normalized \texttt{thenlper/gte-small} embeddings of abstracts and queries to rank eligible candidates by cosine similarity. A submission query combines the agent's research intention and direction keywords. Retrieval returns up to $1{,}000$ unsubmitted candidates from the current calendar year, and the selection prompt displays the top-$k$ with their titles, abstracts, and topics. Citation selection retrieves previously accepted papers in the same world. Exploration and paper-distance metrics use the same embeddings.

\noindent \textbf{Decision inputs and review rubric.}
Research prompts condition on expertise, strategy, resources, and available directions. Paper-selection prompts include the research focus, candidate artifacts, venue topics, and recent paper history. Review prompts include the paper's title, abstract, and topic labels, reviewer expertise, and memory. The rubric requests a $1$--$5$ score and a one-sentence justification (decimals allowed). It also asks reviewers to place more than $60\%$ of papers at $2.0$--$2.5$ or below. Actual acceptance follows within-venue ranking and quotas. Funding evaluation uses applicants' proposals, expertise, paper records, and program priorities.

\noindent \textbf{Decoding and memory.}
Qwen3-32B runs in thinking mode with temperature $0.7$ and structured JSON responses for decision prompts. The default completion budget is $2{,}048$ tokens, with an $8{,}192$-token request for batched funding rankings. Memory is an append-only record of dated events, feedback summaries, and decision rationales, saved with researcher state. When memory is included in a prompt, entries are concatenated in order.

\FloatBarrier
\subsection{Motivation for closed-loop simulation}
\label{app:closed-loop-motivation}
The loop is closed by carrying updated researcher states and scientific records into subsequent decisions.
Rejected manuscripts and their reviews remain available for later resubmission decisions, while accepted papers enter the citable literature and the authors' track records used in funding evaluation.
Funding and action costs update researchers' budgets, determining who can remain active and hence contribute submissions and reviews in later years.
An intervention can therefore change both the outcomes of the current round and the population, resources, and literature on which later rounds operate.

LLMs provide context-dependent choices and evaluations within this loop, including research-direction and paper selection, peer review, and funding assessment.
Explicit institutional rules govern acceptance quotas, resource accounting, and attrition.
This separation allows institutional mechanisms to be varied while retaining the same agent decision framework; the resulting counterfactual trajectories describe their downstream consequences under the specified simulation assumptions.

\begin{figure*}[!htbp]
\centering
\includegraphics[width=0.98\linewidth]{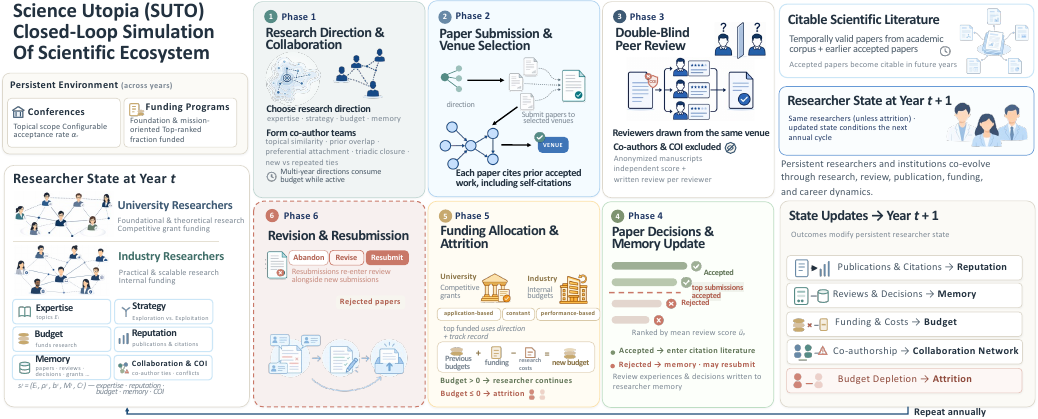}
\vspace{-.1in}
\caption{
\textbf{Overview of \method's LLM-powered research ecosystem.}
% The overall pipeline of Science Utopia where AI is used in all phases.
\method models scientific research as a closed-loop ecosystem where research-direction choice, collaboration, submission, peer review, resubmission, citation, and funding are connected through annual feedback loops, with outcomes updating persistent researcher states and shaping subsequent decisions.
% academic actors and institutional mechanisms co-evolve over simulated years.
% Each simulation world contains interacting components such as researcher agents, publication venues, funding programs, impact trackers, agent memories, and a temporally indexed scientific corpus.
% These components are connected through annual feedback loops: agents choose research directions, form collaborations, submit literature-grounded artifacts, review one another's work, receive publication and funding outcomes, and carry the resulting resources, reputation, memory, citations, and conflicts of interest into later decisions.
}
\label{fig:framework}
\vspace{-3mm}
\end{figure*}

\FloatBarrier
\subsection{Simulation corpus and released dataset}
\label{app:dataset-stats}
% Table~\ref{tab:dataset-stats} summarizes the artifacts generated by the worlds underlying our analyses.
Across the $61$ reported worlds, \method simulates over $40{,}000$ researcher agents affiliated with $8{,}000$ institutions over ${\sim}390{,}000$ researcher-years.
These runs produce ${\sim}177{,}000$ submitted manuscripts, ${\sim}400{,}000$ accept or reject decisions, and ${\sim}1.2$M individual peer reviews with written justifications.
The simulations also record ${\sim}595{,}000$ researcher memory entries, including ${\sim}194{,}000$ research-direction decisions with written rationales. % $285{,}984$ citation links among simulated submissions, and $327{,}367$ funding applications, of which $93{,}356$ receive awards.
We plan to release this corpus, covering per-year snapshots of researcher states, submissions, reviews, decisions, citations, memory records, and funding outcomes, as a resource for research on peer review, research evaluation, and the science of science.

\FloatBarrier
\section{Additional Figures}
\label{app:additional-figures}
\begin{itemize}[leftmargin=*]
    \setlength\itemsep{0em}
    \item Figure~\ref{fig:teaser} is an overview of the annual simulation cycle in \method. Each simulated year proceeds through six phases, and the outcomes of each phase update persistent researcher states that inform decisions in subsequent years.
    \item Figure~\ref{fig:paper_novelty_scores} is the relationship between topical departure and citation impact under the three exploration strategies. Each point is an accepted paper, with topical departure from the author's recent work on the $x$-axis and citations within three years on the $y$-axis; lines show mean citations within departure octiles. Citation gains rise most consistently with departure for cautious explorers, while the relationship is weak for explorers and exploiters.
\end{itemize}

\FloatBarrier
\section{Broader Impact}
\label{app:discussion}
LLMs are increasingly involved throughout scientific workflows, including writing, reviewing, and research decision-making~\citep{kobak2025delving}. Their impact on science, however, may not be captured by evaluating these tasks in isolation. Repeated interactions among researchers, institutions, funding mechanisms, and publication systems can create feedback loops whose long-term consequences only emerge at the ecosystem level.

Our framework provides a controlled testbed for studying these dynamics. By simulating persistent researchers and institutions under explicit rules, it enables counterfactual experiments on mechanisms that are difficult to isolate in real scientific systems, revealing how seemingly local changes---to review criteria, funding incentives, research strategies, or AI-mediated behavior---can accumulate into system-level changes in workload, resource allocation, scientific impact, and diversity. More broadly, we view such ecosystem simulation as a complementary tool for anticipating how increasingly AI-mediated science may evolve and for studying the institutional mechanisms that shape its long-term trajectory.

\end{document}